\documentclass{article}
\usepackage{graphicx}

\usepackage[nonatbib,preprint]{neurips_2020}
\usepackage[square,sort&compress,numbers]{natbib}
\usepackage[utf8]{inputenc} 
\usepackage[T1]{fontenc}    
\usepackage{hyperref}       
\usepackage{url}            
\usepackage{booktabs, tabu}       
\usepackage{amsfonts}       
\usepackage{nicefrac}       
\usepackage{microtype}      
\usepackage{amsmath,amssymb}
\usepackage{algorithm}
\usepackage{wrapfig}
\usepackage[noend]{algpseudocode}
\usepackage{caption}
\usepackage{tikz}
\usepackage{bm,braket}
\usepackage[T1]{fontenc}
\usepackage{bm}
\usepackage[font=small,labelfont=bf,tableposition=top]{caption}
\usepackage{titling}

\DeclareCaptionLabelFormat{andtable}{#1~#2  \&  \tablename~\thetable}

\definecolor{mydarkblue}{rgb}{0,0.08,0.45}
\definecolor{myfavblue}{rgb}{0.1176, 0.392, 1.0}
\hypersetup{
    colorlinks=true,
    linkcolor=mydarkblue,
    citecolor=mydarkblue,
    filecolor=mydarkblue,
    urlcolor=mydarkblue}

\newcommand{\beginsupplement}{%
        \setcounter{table}{0}
        \setcounter{figure}{0}
        \setcounter{section}{0}
		\renewcommand\thesection{\Alph{section}}
     }

\makeatletter
\newcommand*\bigcdot{\mathpalette\bigcdot@{.5}}
\newcommand*\bigcdot@[2]{\mathbin{\vcenter{\hbox{\scalebox{#2}{$\m@th#1\bullet$}}}}}
\makeatother
 
\usepackage{url}

\title{Molecular machine learning with conformer ensembles}

\author{
  Simon Axelrod$^{\star, \dagger}$ \  and Rafael G\'{o}mez-Bombarelli$^{\dagger}$  \\
  $^{\star}$ Department of Chemistry and Chemical Biology\\
  Harvard University \\
  Cambridge, MA, 02138 \\
  $^{\dagger}$ Department of Materials Science and Engineering \\
  Massachusetts Institute of Technology \\
  Cambridge, MA, 02142 \\ 
  \texttt{simonaxelrod@g.harvard.edu, rafagb@.mit.edu} 
}

\begin{document}
\maketitle

\begin{abstract}

Virtual screening can accelerate drug discovery by identifying promising candidates for experimental evaluation. Machine learning is a powerful method for screening, as it can learn complex structure-property relationships from experimental data and make rapid predictions over virtual libraries. Molecules inherently exist as a three-dimensional ensemble and their biological action typically occurs through supramolecular recognition. However, most deep learning approaches to molecular property prediction use a 2D graph representation as input, and in some cases a single 3D conformation. Here we investigate how the 3D information of multiple conformers, traditionally known as 4D information in the cheminformatics community, can improve molecular property prediction in deep learning models. We introduce multiple deep learning models that expand upon key architectures such as ChemProp and Schnet, adding elements such as multiple-conformer inputs and conformer attention. We then benchmark the performance trade-offs of these models on 2D, 3D and 4D representations in the prediction of drug activity using a large training set of geometrically resolved molecules. The new architectures perform significantly better than 2D models, but their performance is often just as strong with a single conformer as with many. We also find that 4D deep learning models learn interpretable attention weights for each conformer.

\end{abstract}

\section{Introduction}

Drug development is a long and costly process. Bringing a new drug to market takes an average of seven years \cite{drug_time} and costs \$2.9 billion (USD, 2013) \cite{drug_cost}. Because so many drugs fail in late stage trials, it is critical to generate a variety of leads to increase the chance of success. Leads are often discovered by screening large chemical libraries, but these libraries are expensive to manage, and their chemistry is both homogeneous and poorly reflective of actual drug chemistry \cite{brown2014trends}. Computational screening can improve this process by exploring a much larger space of compounds and identifying the top candidates for experimental testing. Such methods range from physics-based simulations, such as computational docking \cite{trott2010autodock, li2006tarfisdock} and molecular dynamics \cite{alonso2006combining}, to data-driven regressions, such as machine learning (ML) \cite{burbidge2001drug, vamathevan2019applications, ecoli_1}. Combinations of both have also produced fruitful results \cite{shen2020machine}. While docking accuracy is limited by the scoring function and force field \cite{trott2010autodock}, the accuracy of ML methods is mostly limited by the amount of available data, and ML inference is orders of magnitude faster.

Given enough data, neural networks used in ML can in principle learn any function, including the mapping from molecule to property. However, there is typically a scarcity of data for successful drugs, as the vast majority of tested molecules do not bind the target protein. It is therefore necessary to optimize the neural network architecture to best leverage the limited data. Advances in deep learning have improved ML performance by training networks directly on molecular graphs \cite{bartok2013representing, duvenaud_convolutional_2015, li2015gated, battaglia2016interaction, kearnes2016molecular, schutt2017quantum, schnet_1, schnet_2, Klicpera2020, thomas2018tensor, feinberg2018potentialnet, unke2019physnet, liu2020transferable}. These message-passing neural networks (MPNNs) use graphs to generate learnable, continuous fingerprints, which are then used as input to a neural network that outputs predicted properties. This representation encodes a task-specific molecular similarity, as fingerprints predicted to have similar properties are themselves similar. 

The state of the art in deep learning uses 2D molecular graphs \cite{chemprop} (or in some cases a single 3D molecular structure \cite{bartok2013representing}) to generate these fingerprints. One way to further improve the representation is to use 3D ensemble information. A molecule is neither a 2D structure nor a single 3D structure, but rather an ensemble of continuously inter-converting 3D structures. The process of drug binding is a 3D recognition event between the drug and the binding pocket, and so depends critically on the 3D structures accessible to the molecule. Ensembles have been used to generate fixed descriptors in computational drug discovery \cite{andrade2009rational}, but they have not been used in MPNNs. Nevertheless, the ensemble information may still be learned implicitly by 2D MPNNs, as the set of conformers is in principle a function of the graph only. The extent to which 2D models can implicitly learn this information, and thus match the performance of single- or multi-conformer 3D models, is not currently understood.

\begin{figure}[t]
    \centering
    \includegraphics[width=0.5\textwidth]{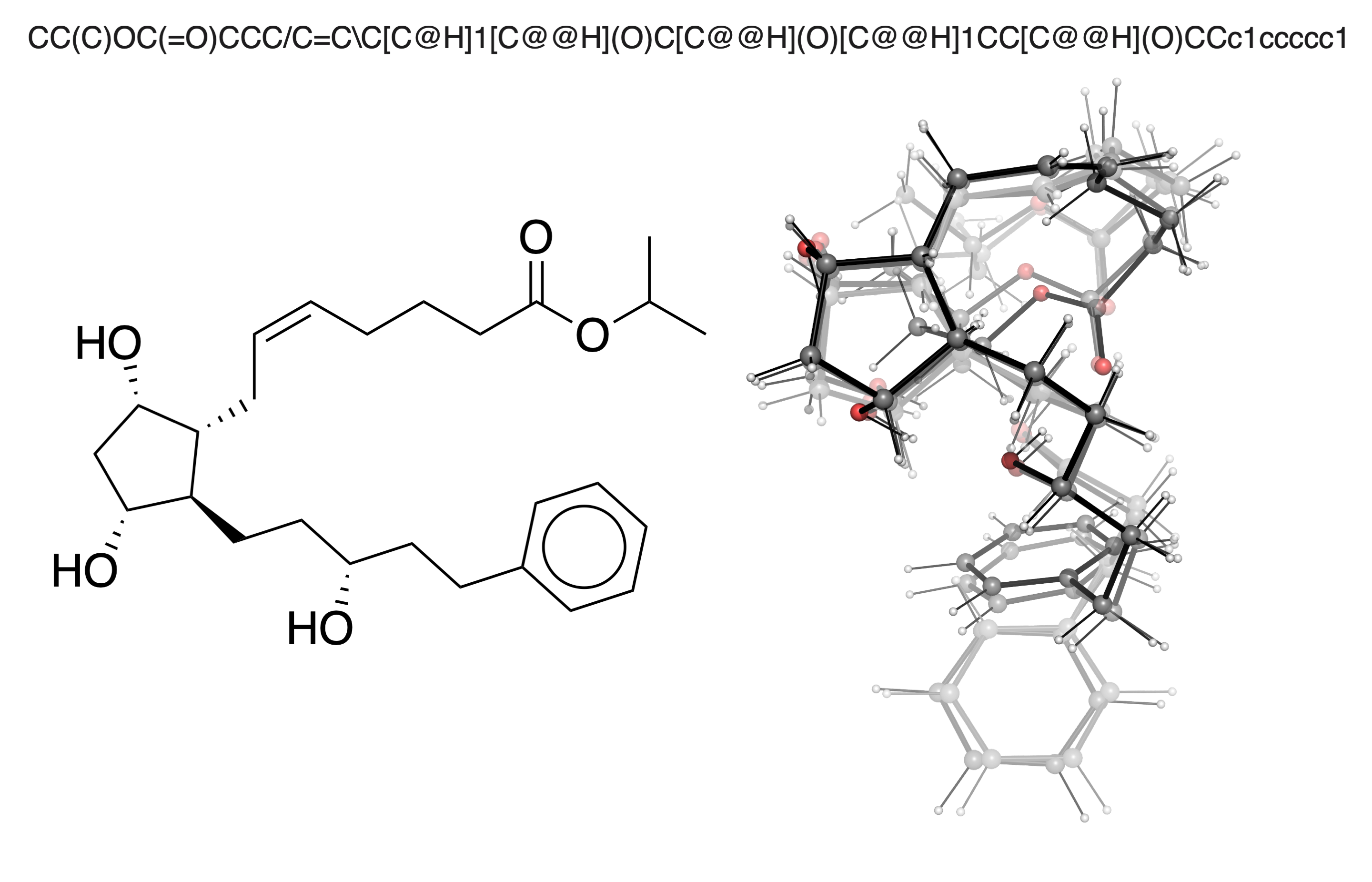}
    \caption{Molecular representations of the latanoprost molecule. \textit{top} SMILES string. \textit{left} Stereochemical formula with edge features, including wedges for in- and out-of-plane bonds, and a double line for \textit{cis} isomerism. \textit{right} Overlay of conformers. Higher transparency corresponds to lower statistical weight. }
    \label{fig:reps}
\end{figure}

Here we investigate whether 3D information of one or more conformers can improve computational drug discovery with ML (Fig. \ref{fig:reps}). As an example we screen molecules that can inhibit the novel SARS coronavirus 2 (SARS-CoV-2), which causes COVID-19 \cite{covid_id}. 
We find that models based on a single conformer can better identify held out CoV-2 inhibitors than models based on 2D graphs. However, using multiple conformers does not further improve the results. We also test a transfer learning strategy to leverage the large amount of data for SARS-CoV, which causes SARS, to better predict SARS-CoV-2 inhibition. We find that transfer learning with 3D and 4D models can be better than training 2D models from scratch, but that the improvement is small. 


\section{Previous work}
We distinguish two broad approaches to computational drug discovery. The first, which we refer to as QSAR or classical QSAR (quantitative structure-activity relationship), has long been applied by the computational chemistry community. The second, which we refer to as deep learning, has seen growing interest in the last years driven by advances in representation learning, such as graph convolutional neural networks. Both approaches predict molecular properties by applying a readout function to a molecule's features. Readout methods include linear, inherently regularized functions like partial least-squares \cite{verma20103d}, and nonlinear, flexible functions like $k$-nearest neighbors \cite{ajmani2006three} and neural networks. However, MPNNs differ from classical QSAR approaches, including those with neural network readouts, in that they learn the features directly from the molecules in the training set. 

Both the cheminformatics and deep learning communities have focused on the improvement of so-called featurization or representation learning. Researchers in computational drug discovery have developed features that better reflect the molecule-protein binding process (see below). Machine learning researchers have developed models that better ``train themselves'' to learn powerful featurizations. MPNNs can generate strong molecular representations that outperform hand-crafted descriptors, but require vast amounts of data to do so. In practice their performance is often improved by incorporating features developed by experts \cite{chemprop}. Here we review various 3D and 4D methods that have been extensively developed by the drug discovery community and successfully applied to many targets \cite{kim1998list, ortuso2006gbpm, holzgrabe1996conformational, rhyu19953d, tokarski1994three, cardozo1992qsar, magdziarz2009receptor, gieleciak2007modeling, magdziarz20063d, niedbala2006comparative, jojart20053d, rush2005shape, senese20044d, iyer2007treating, romeiro2005construction, liu20034d, pasqualoto2004rational, hong20033d, krasowski20024d, thipnate20093d, ravi20014d}. We then briefly review different 2D and 3D approaches in deep learning and discuss how the principles of 4D QSAR can be applied to MPNNs.

\subsection{3D QSAR}
Classical QSAR methods generate features based on physical principles. One such principle is that binding affinity and specificity are determined by the forces between ligands and receptors. 
For example, the popular CoMFA method \cite{cramer1988comparative, podlogar2000qsar} aligns molecules in a 3D grid, computes steric and electrostatic fields at different positions around the molecules, combines these values into features, and correlates the features with biological activity using partial least-squares \cite{verma20103d}. The GRID method \cite{goodford1985computational, kim1998critical, kim2001thermodynamic} also incorporates hydrophobic and hydrogen bonding interactions.

Other methods use the principle that ligands with similar shapes have similar binding properties. For example, the MSA method uses the difference in steric volume between a sampled structure and a reference structure to predict the sample's activity \cite{hopfinger1980qsar}. The reference structure is chosen as the geometry in the training set that maximizes the quality of the fit. Other methods, such as ROCS \cite{rush2005shape, hawkins2007comparison} and Phase Shape \cite{sastry2011rapid}, follow a similar principle but use different methods for computing the volume overlap. 

In all 3D methods one must choose a molecular conformation for each species. Molecules exist as an ensemble of conformers, but the most important is the one with the strongest binding affinity to the target. This is known as the bioactive conformation. Many 3D methods assume that the lowest energy conformer is the bioactive conformation \cite{verma20103d, oprea20033d}, but this is often not the case. Others, such as Compass \cite{jain1994compass}, generate a model and an estimate of the bioactive pose together. 


\subsection{4D QSAR}
4D methods use multiple conformations for each molecule. For example, the method of Ref. \cite{hopfinger1997construction} positions each conformer in a 3D grid, assigns atom features to occupied grid sites, and averages over all conformers. Refs. \cite{albuquerque1998four, albuquerque2007multidimensional} use a similar approach but with frames generated by MD simulations. As in 3D models, one can identify the bioactive pose of a new species by selecting the conformer with the highest predicted activity. Even though the binding pocket is not used during training, the interactions between the predicted bioactive pose and the binding site tend to be energetically favorable \cite{andrade2009rational}.

Ref. \cite{ash2017characterizing} used molecular dynamics trajectories of the ligand-protein complex to generate an ensemble of WHIM shape descriptors \cite{todeschini2003descriptors}, and used the mean and standard deviation of the descriptors as additional features. WHIM descriptors are invariant to translation and rotation, while grid-based and volume overlap methods depend on the orientation of the molecule. Therefore, the alignment of each molecule to a reference structure, which is often ambiguous and can significantly alter model predictions \cite{akamatsu2002current, kim1995comparative, verma20103d}, is avoided. Further, explicitly identifying a bioactive conformation was not necessary because the MD simulations were performed with the full ligand-receptor complex. The MD descriptors were shown to strongly distinguish between the most active and the moderately/weakly active and inactive ERK2 kinase inhibitors. They were also shown to provide information absent in 2D descriptors, and even in 3D descriptors from binding poses generated with computational docking. However, since the approach used the protein-ligand complex in MD simulations, it is not clear whether the results would also apply to ligand-only calculations. 

\subsection{MPNNs}

Many QSAR methods generate features based on physical principles that are relevant to the target property. A conceptually different approach is to generate fingerprints that encode the maximum amount of molecular information and can be applied in any context. 
The current state of the art in deep learning uses MPNNs \cite{bartok2013representing} to learn molecular representations directly from 2D or 3D molecular graphs \cite{duvenaud_convolutional_2015, li2015gated, battaglia2016interaction, kearnes2016molecular, schutt2017quantum, schnet_1, schnet_2, Klicpera2020, thomas2018tensor}. 

\subsection{QSAR and MPNNs}
Extensive work has been done in the deep learning community to develop and improve MPNNs that act on 2D or single 3D molecular graphs \cite{bartok2013representing, duvenaud_convolutional_2015, li2015gated, battaglia2016interaction, kearnes2016molecular, schutt2017quantum, schnet_1, schnet_2, Klicpera2020, thomas2018tensor}, but no work has been done on representations with ensembles. Moreover, unlike in QSAR, the issue of bioactive conformation has not been addressed in deep learning, as the lowest energy conformer is the only structure used, and this may not be the bioactive conformation.

Another issue for 3D and 4D models is the generation of conformers. Conformer generation is a difficult task (see Subsection \ref{subsec:conf} below), and the ranking of conformers by energy with classical force fields is highly inaccurate \cite{Kanal2018}. MD simulations may be able to better sample low-energy regions of phase space than stochastic conformer generators, but without advanced sampling methods can become stuck in local minima. Moreover, the quality of the classical MD conformers is still limited by the accuracy of the force field. This problem affects all models that use single or multiple 3D structures, whether the featurization is hand-crafted or learned.

In this work we combine the physical picture of a bioactive pose, used extensively in QSAR, with tools developed in deep learning. First we develop a number of new 3D MPNN models by combining force field architectures with 2D property predictors. Next we show how the MPNNs can be trained on multiple conformers or on a single ``effective conformer'' that represents the ensemble. The former uses the deep learning concept of attention \cite{bahdanau2014neural, kim2017structured, noam_lr, gat} to learn the bioactive pose concurrently with the model training. We also address the issue of inaccurate conformer generation, common to all 3D and 4D models, by using the high-quality conformers in the GEOM dataset \cite{axelrod2020geom}. The GEOM conformers were generated with the CREST method \cite{crest}, which uses semi-empirical DFT and advanced sampling to thoroughly and accurately sample the low-energy regions of phase space. Our work thus expands upon methods used in 2D and 3D deep learning, addresses conformer issues common to all 3D and 4D methods, and provides a number of ways to add 4D information to MPNNs.


\section{Methods}

\subsection{Neural network architectures}

Here we discuss and extend various 2D and 3D message-passing architectures, and describe several methods for applying them to conformer ensembles.

\subsubsection{Message passing}
Molecules can be represented as graphs. These graphs consist of a set of nodes (atoms) connected to each other by a set of edges. Both the nodes and edges have features. The atoms, for example, can be characterized by their atomic number and partial charge. The edges can be characterized by bond type or by interatomic distance. In the \textit{message passing phase}, message-passing neural networks (MPNNs) aggregate node and edge features to create a learned fingerprint. The \textit{readout phase} uses the fingerprint as input to a regressor that predicts a property \cite{chemprop}.

The message passing phase consists of $T$ steps, or convolutions. In what follows, superscripts denote the convolution number. The node features of the $v^{\mathrm{th}}$ node are $\vec{x}_v$, and the edge features between nodes $v$ and $w$ are $\vec{e}_{vw}$. The atom features $\vec{x}_v$ are first mapped to so-called hidden states $\vec{h}_v^{0}$. A message is $\vec{m}_v^{t+1}$ is created in the $t^{\mathrm{th}}$ convolution, which combines $\vec{h}_v$ and $\vec{h}_w$ for each pair of nodes $v$ and $w$ with edge features $\vec{e}_{vw}$ \cite{mpnn, chemprop}:
\begin{align}
\vec{m}_v^{t+1} = \sum_{w \in N(v)} M_t(\vec{h}_v^t, \vec{h}_w^t, \vec{e}_{vw}),
\end{align}
where $N(v)$ is the set of neighbors of $v$ in graph $G$, and $M_t$ is a message function. The hidden states are updated using a vertex update function $U_t$:
\begin{align}
\vec{h}_v^{t+1} = U_t(\vec{h}_v^t, \vec{m}_v^{t+1}).
\end{align}
The readout phase then uses a function $R$ to map the final hidden states to a property $y$, through
\begin{align}
\hat{y} = R({\vec{h}_v^T  | v \in G}).
\end{align}

In this work we also use the \textit{directed} message-passing ChemProp model \cite{chemprop}, which achieves state-of-the-art performance on a wide range of prediction tasks \cite{chemprop}. In this implementation, hidden states $\vec{h}_{vw}^{  t}$ and messages $\vec{m}_{vw}^t$ are used, rather than node-based states $\vec{h}_v^{  t}$ and messages $\vec{m}_v^t$. Hidden states are initialized with
\begin{align}
\vec{h}^{  0 }_{vw} = \tau(\mathbf{W}_i \ [\vec{x}_v ||  \vec{e}_{vw}]),
\end{align}
where $\mathbf{W}_i \in \mathbb{R}^{F \times F_0}$ is a learned matrix, $F$ is the hidden state dimension, $F_0$ is the dimension of $[\vec{x}_v || \vec{e}_{vw}] $, which is the concatenation of the atom features $\vec{x}_v$ for atom $v$ and the bond features $\vec{e}_{vw}$ for bond $vw$, and $\tau$ is an activation function. The message passing function is simply $M_t(\vec{x}_v, \vec{x}_w, \vec{h}^{  t}_{vw}) = \vec{h}^{  t}_{vw}$. The edge update function is the same neural network at each step:
\begin{align}
U_t(\vec{h}_{vw}^{  t}, \vec{m}_{vw}^{t+1} ) = U(\vec{h}_{vw}^{  t}, \vec{m}_{vw}^{t+1}  ) = \tau(\vec{h}_{vw}^{  0} + \mathbf{W}_m \vec{m}_{vw}^{t+1}  ), 
\end{align}
where $\mathbf{W}_m \in \mathbb{R}^{F \times F}$ is a learned matrix. Each message-passing phase is then 
\begin{align}
& \vec{m}_{vw}^{t+1} = \sum_{k \in N(v) \text{\textbackslash} w } \vec{h}_{kv}^{  t} , \ \ \ \ 
 \vec{h}_{vw}^{  t+1} = \tau ( \vec{h}_{vw}^0 + \mathbf{W}_m \vec{m}_{vw}^{t+1} ),
\end{align}
for $t \in {1, ..., T}$. After the final convolution, the atom representation of the molecule is recovered through
\begin{align}
& \vec{m}_v = \sum_{w \in N(v)} \vec{h}_{vw}^{  T}, \ \ \ 
 \vec{h}_v = \tau(\mathbf{W}_a [\vec{x}_v || \vec{m}_v] ). \label{eq:edge_to_node}
\end{align}
The hidden states are then summed to give a feature vector for the molecule: $\vec{h} = \sum_{v \in G} \vec{h}_v$. Properties are predicted through $\hat{y} = f(\vec{h})$, where $f$ is a feed-forward neural network. In ChemProp the atom features are atom type, number of bonds, formal charge, chirality, number of bonded hydrogen atoms, hybridization, aromaticity, and atomic mass. The bond features are the bond type (single, double, triple, or aromatic), whether the bond is conjugated, whether it is part of a ring, and whether it contains stereochemistry (none, any, E/Z or cis/trans). All features are one-hot encodings. The ChemProp code was accessed through \cite{chemprop_git}.

\subsubsection{Learning with 3D features}
A variety of graph convolutional models have been proposed for learning force fields, which map a set of 3D atomic positions of a molecular entity to an energy. Architectures designed for force fields typically do not incorporate information about the covalent connectivity \cite{schnet_1, schnet_2, Smith2017, Smith2017b} since these bonds are broken and formed during chemical reactions and may not be clearly defined. This differs from architectures for property prediction, which are typically based on 2D graphs \cite{duvenaud_convolutional_2015, chemprop, dai} but can also use 3D information \cite{mpnn}. 

Here we combine graph-based models with 3D models in a number of ways. Our first such model is called \textbf{SchNetFeatures}, as it uses the \textbf{SchNet} force field architecture \cite{schnet_1, schnet_2} (code adapted from \cite{schnet_pack}), but adds additional graph-based \textbf{features} based on the covalent connectivity. Our second model is called \textbf{ChemProp3D}, as it uses the \textbf{ChemProp} property prediction architecture, but adds additional \textbf{3D} (distance-based) edge features between atoms within a distance threshold. We have also explored a reduced ChemProp3D model, in which bond states are updated only based on other covalently-bonded neighbors, and not on non-bonded neighbors. In this case the updated bond features are simply concatenated with the original distance features at each step. This model is called \textbf{CP3D-NDU}, for \textbf{C}hem\textbf{P}rop\textbf{3D} with \textbf{n}o \textbf{d}istance \textbf{u}pdates. We also abbreviate this as CND.

In the typical SchNet model the feature vector of each atom is initialized with an embedding function. This embedding generates a random vector that is unique to every atom with a given atomic number, and is also learnable. The edge features at each step $t$ are generated through a so-called filter network $V^t$. The filter network generates an edge vector $\vec{e}_{vw}$ by expanding the distance $\vert \vert \vec{r}_v - \vec{r}_w \vert \vert$ in a basis of Gaussian functions. The expansion coefficients are then transformed into $\vec{e}_{vw}$ through linear and non-linear operations. Because only the distance between two atoms is used to create $\vec{e}_{vw}$, the features produced are invariant to rotations and translations.

In each convolution $t+1$, the new messages and hidden vectors are given by
\begin{align}
& \vec{m}_v^{t+1} = \sum_{w} M_t(\vec{h}_v^t, \vec{h}_w^t, \vec{e}_{vw}) = \sum_{w} \vec{h}_v^t \circ \vec{e}_{vw}, \nonumber \\
& \vec{h}_v^{t+1} = U_t(\vec{h}^t_v, \vec{m}_v^{t+1}) = \vec{h}_v^t 
+ I^t\left(\sum_{w \in N(v)} J^t(\vec{h}_w^t) \circ \vec{e}_{vw} \right),
 \label{eq:schnet}
\end{align} 
where the edge features are 
\begin{align}
& \vec{e}_{vw} = V^t(r_{vw} ).
\end{align}
Here $\circ$ denotes element-wise multiplication, and $I^t$ and $J^t$ are a series of linear and non-linear operations applied to the atomic features. The neighbors $N(v)$ of atom $v$ are those within a pre-set cutoff distance $r_{\mathrm{cut}}$. For SchNetFeatures we replace the initial atomic feature embedding with the graph-based atom features described above for ChemProp. We also combine distances features with bond features to create the $\vec{e}_{vw}$ of Eq. (\ref{eq:schnet}), through
\begin{align}
    & \vec{e}_{vw} \to [\vec{e}_{vw}^{ \ \mathrm{dist}} \ || \ \vec{e}_{vw}^{ \ \mathrm{bond}}] = [V^t(r_{vw}) \ || \ \vec{e}_{vw}^{ \ \mathrm{bond}}]  \label{eq:schnet_feat}.
\end{align}
Here $\vec{e}_{vw}^{ \ \mathrm{dist}} $ are the edge features generated by the SchNet distance filter network. The $\vec{e}_{vw}^{ \ \mathrm{bond}}$ are hidden bond vectors, obtained from a nonlinear operation applied to the bond features, and are set to 0 for non-bonded atoms pairs. Note that applying $I^t$ in Eq. (\ref{eq:schnet}) mixes the distance and bond features before they are used to update the $h^t_v$. 

In the original SchNet implementation the readout layer converted each atomic feature vector into a single number, and the numbers were summed to give an energy. Consistent with the the notion of property prediction, we here instead convert the node features into a molecular fingerprint by adding the features of each node. The readout function is then applied to this fingerprint. 

The usual ChemProp model creates and updates edge embeddings for each bond. For ChemProp3D we create an edge embedding for each pair of atoms separated by less than $r_{\mathrm{cut}}$. In particular, we use one initialization matrix for the distances, $\mathbf{W}_{i}^{\mathrm{dist}}$, and one for the bond features, $\mathbf{W}_{i}^{\mathrm{bond}}$. Hidden states are then initialized through
\begin{align}
    & \vec{h}_{vw}^{0, \ \mathrm{dist}} =  \tau(\mathbf{W}_i^{\mathrm{dist}} \ [\vec{x}_v \ || \  \vec{e}^{ \ \mathrm{dist}}_{vw}] ) \nonumber \\
    & \vec{h}_{vw}^{0, \ \mathrm{bond}} =  \tau(\mathbf{W}_i^{\mathrm{bond}} \ [\vec{x}_v \ || \  \vec{e}^{ \ \mathrm{bond}}_{vw}] ) \nonumber \\
    & \vec{h}_{vw}^{0} = [\vec{h}_{vw}^{ 0, \  \mathrm{dist}} \ || \  \vec{h}_{vw}^{ 0, \  \mathrm{bond}}].
\end{align}
The distance features $e^{\mathrm{dist}}_{vw}$ are the result of a SchNet filter network applied to the distances $r_{vw}$. The bond features for non-bonded pairs are again set to 0, and the remainder of the ChemProp architecture is unchanged (apart from the neighbors now including all atom pairs within $r_{\mathrm{cut}}$ of each other). 

The CND model differs from ChemProp3D in that only the edge features of bonded pairs are updated. These updates do not use any distance information, meaning that they are equivalent to the updates in the 2D ChemProp model. Distance information is instead incorporated in the following way. After all $T$ convolutions are complete, a set of distance features $\vec{e}_{vw}^{ \ \mathrm{dist}}$ is generated for all pairs $vw$ within $r_{\mathrm{cut}}$ of each other. The $\vec{e}_{vw}^{ \ \mathrm{dist}}$ of bonded pairs are concatenated with the final bonded edge features $\vec{m}_{vw}^{T}$, while those of non-bonded pairs are concatenated with zeros. These concatenated vectors are then summed as in Eq. (\ref{eq:edge_to_node}) to give the node features $\vec{m}_v$, which are in turn summed to give the molecular features $h$.

For all 3D models we use 10 Gaussian functions for each distance and a cutoff distance of 5 \AA.  We use three convolutions for SchNetFeatures, but use two convolutions for the ChemProp3D models because of higher memory costs. The SchNet force field uses Gaussians spaced 0.1 \AA \ apart \cite{schnet_1, schnet_2}, which is necessary to resolve movements that lead to appreciable changes in energy. Here we use a much larger spacing (0.3-0.5 \AA) because we are interested in larger-scale features of the molecule. Resolving the structure at a finer scale would not be expected to improve performance, and could more easily lead to overfitting. 

The above discussion applies to molecules associated with one geometry. However, multiple conformers correspond to a single stereochemical formula and can be considered at the same time, each one having a different statistical weight $p^{(n)}$. It is not immediately clear how to pool the fingerprints of the different conformers. A simple pooling scheme would be to multiply the fingerprint of each conformer by $p^{(n)}$ and add the results. However, this assumes that the resultant property is a statistical average of individual conformer properties, and this is not always the case. For example, the contribution of a conformer to the binding process is not determined by its statistical weight. Rather, it is determined by the affinity of the conformer for the target, but this affinity is not known \textit{a priori}, and large differences in binding energy may compensate for small differences in conformational energy. 

\subsection{Attention model}
An ideal pooling scheme would be adaptable, learning fingerprint weights that are best suited to the task at hand. For this reason we propose a pooling mechanism based on attention \cite{bahdanau2014neural, kim2017structured, noam_lr, gat}. Attention combines a learned vector with two feature vectors to determine the importance of one set of features to another. The resulting coefficients are then normalized to give weights for each feature vector. Similarly to other applications of attention \cite{gat}, we have found that it is useful to include multiple attention heads in the pooling. These different heads can learn different features to focus on in the pooling. 

The notion of attention can be applied to conformer fingerprints in one of two ways. The first is to let the attention mechanism learn the importance of one conformer's features to another, giving a set of attention coefficients $\alpha_{nm}$ for $n, m \in N$, where $N$ is the number of conformers. The second, which we call linear attention, is to simply learn the conformer's overall importance, rather than its importance with respect to another conformer. This gives a set of coefficients $\alpha_{n}$ for each conformer. The choice of attention or linear attention is treated as a hyperparameter to be optimized.

\subsubsection{Incorporating statistical weights}
The attention mechanism should be able to use the statistical weight of each conformer if it improves predictions. However, the fingerprint of the $n^{\mathrm{th}}$ conformer, $\vec{h}^{(n)}$, does not take into account $p^{(n)}$. To add this information we embed the weight as a vector $\vec{d}^{ \ (n)}$, through
\begin{align}
\vec{d}^{ \ (n)} = \mathrm{SoftMax} \big( \mathbf{D} p^{(n)} + \vec{b} \big), 
\end{align}
where $\mathbf{D} \in \mathbb{R}^{S \times 1} $ is a learned matrix, $S$ is the dimension of $\vec{d}^{ \ (n)}$, and $\vec{b}$ is a learned bias. The softmax activation means that $\vec{d}^{ \ (n)}_k$ can be interpreted as the projection of $p^{(n)}$ onto the $k^{\mathrm{th}}$  probability bin. Here we choose $S=10$ so that the probabilities are divided into 10 bins. A linear layer is then applied to the concatenation of $\vec{d}^{ \ (n)}$ with $\vec{h}^{(n)}$, yielding the final conformer fingerprint $\vec{q}^{ \ (n)}$:
\begin{align}
    \vec{q}^{ \ (n)} = \mathbf{H} \ [ \vec{h}^{(n)} || \vec{d}^{ \ (n)} ] + \vec{b}.
\end{align}
Here, $\mathbf{H} \in \mathbb{R}^{F \times (F+S)} $ is a learned matrix, and $\vec{b}$ is again a learned bias. 
\subsubsection{Computing attention coefficients}
\begin{figure}[t]{}
    \centering
    \vspace{-0.5cm}
    \begin{tikzpicture}
    \node[inner sep=0pt] at (-0.2,0)
        {\includegraphics[width=0.95\textwidth]{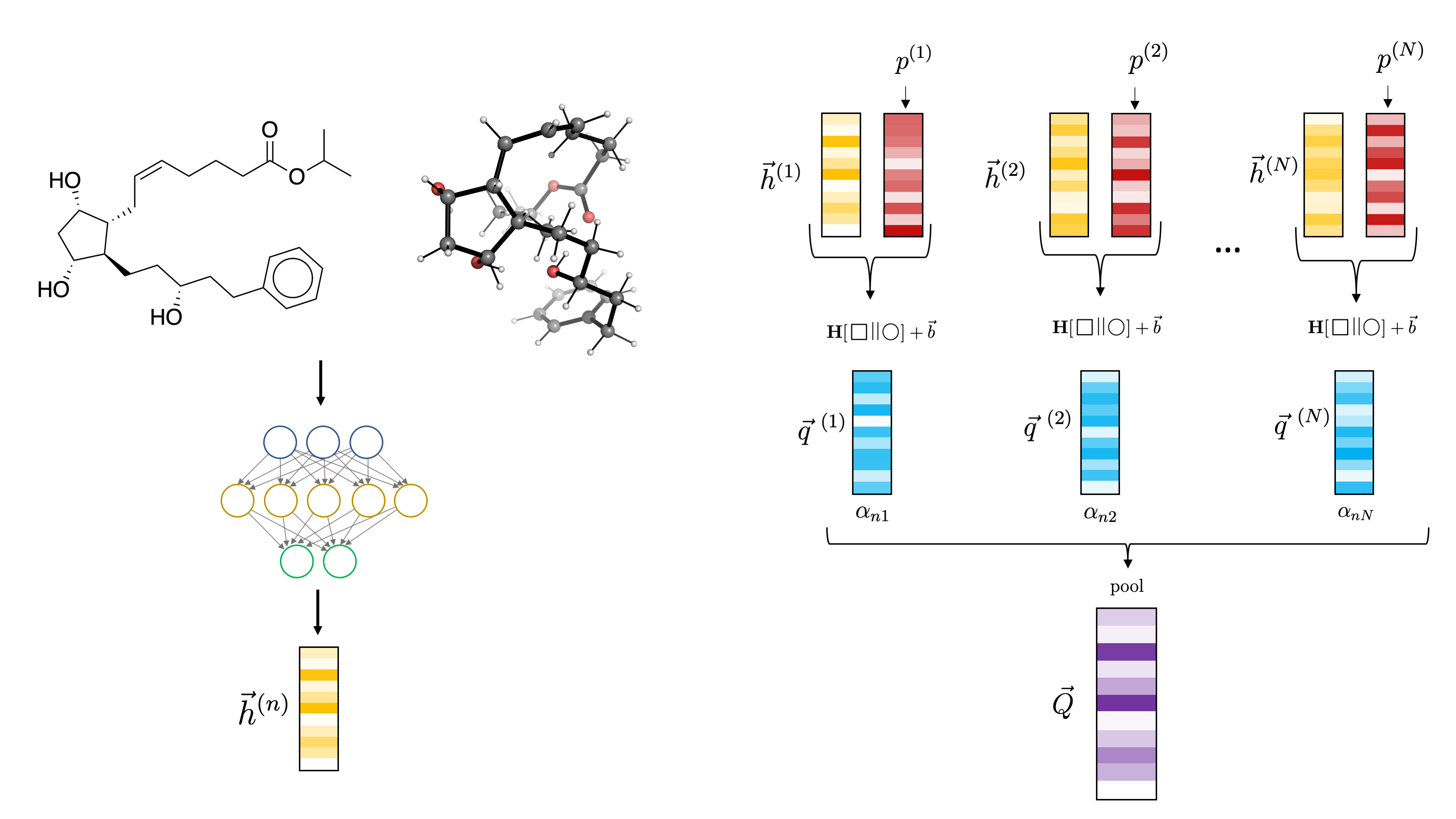}};
    \node[inner sep=0pt] at (-7.5,2.7){(a)};
    \node[inner sep=0pt] at (-0.35,2.7){(b)};
    \end{tikzpicture}
    \caption{Schematic of the architecture in this work. (a) A neural network uses the 3D structure of the $n^{\mathrm{th}}$ conformer, along with its graph information, to create a molecular fingerprint $\vec{h}^{(n)}$. (b) Each fingerprint is combined with an embedding of the conformer's statistical weight $p^{(n)}$. The new fingerprints $\vec{q}^{ \ (n)}$ are aggregated through an attention mechanism to yield the final fingerprint $\vec{Q}$. }
    \label{fig:schematic}
\end{figure}

For the pairwise attention method we then compute the attention coefficients $c_{nm}$ between conformers $n$ and $m$ as
\begin{align}
c_{nm} = \vec{a} \bigcdot [\mathbf{A} \vec{q}^{ \ (n)} || \mathbf{A} \vec{q}^{ \ (m)}],
\end{align}
where $\vec{a} \in \mathbb{R}^{2F}$ is a vector, $\mathbf{A} \in \mathbb{R}^{F \times F}$ is a learned matrix, and $\bigcdot$ is the dot product. For ease of notation we have supressed the dependence on the attention head index. The attention paid by conformer $n$ to conformer $m$ is then
\begin{align}
\alpha_{nm} = \frac{ \mathrm{exp} \big( \mathrm{LeakyReLU} \big( c_{nm} \big) \big)}{\sum_{l} \mathrm{exp} \big( \mathrm{LeakyReLU} \big( c_{nl} \big) \big)}.
\end{align}
The pooled fingerprint $\vec{Q}$ is a weighted sum over each fingerprint followed by an activation function $\tau$, such that
\begin{align}
\vec{Q} = \tau\left( \frac{1}{N} \sum_{nm} \alpha_{nm} \ \mathbf{A} \vec{q}^{ \ (m)} \right).
\end{align}
Here $N$ is the number of conformers, and the division by $N$ ensures that $\sum_{nm} \alpha_{nm} / N = 1$. For the linear attention mechanism only one conformer is used to determine each coefficient, through $c_{n} = \vec{a} \bigcdot (\mathbf{A} \vec{q}^{ \ (n)})$, where $\vec{a} \in \mathbb{R}^{F}$. The coefficients $\alpha_n$ are normalized by summing over one index only, and the pooled conformers are
\begin{align}
\vec{Q}_{\mathrm{linear}} = \tau\left( \sum_{n} \alpha_{n} \ \mathbf{A} \vec{q}^{ \ (n)} \right).
\end{align}
Adding a superscript to denote fingerprints from different heads, the final fingerprint resulting from $K$ linear or pair-wise attention heads is then
\begin{align}
\vec{Q} = [\vec{Q}^{(1)}  \ || \ \vec{Q} ^{(2)}  \ || \ ... \ || \ \vec{Q}^{(K)}].
\end{align}
A schematic of the fingerprinting architecture is shown in Fig. \ref{fig:schematic}.

\subsection{Average distance model}
As an alternative to the attention-based model we also propose an average-distance or average neighbors (``avg nbrs'') model. In this approach, message-passing occurs over a single effective conformer, with interatomic distances given by the distances averaged over all conformers. That is, the effective distance $\bar{d}_{ij}$ between atoms $i$ and $j$ is given by
\begin{align}
\bar{d}_{ij} = \sum_n w_n d_{ij}^{(n)}, \label{eq:d_bar}
\end{align}
where the sum is over all conformers, $w_n$ is the statistical weight of the $n^{\mathrm{th}}$ conformer, and $d_{ij}^{(n)}$ is the distance between atoms $i$ and $j$ in the $n^{\mathrm{th}}$ conformer. The benefit of this approach is the enormous reduction in computational cost compared to using multiple conformers. Since only one effective graph is needed per species, the computational cost is reduced by a factor of $n_{\mathrm{confs}}$, the average number of conformers per species. The drawback, however, is that we are forced to use non-learnable weights, as learnable weights would require each conformer to be stored in the computation graph. Here we have used statistical weights, but it would also be possible to use any other weights that are determined before training. For example, one could first train an attention model on non-learnable fingerprints, and then use the resulting weights in Eq. (\ref{eq:d_bar}). 

It is important to notice that Eq. (\ref{eq:d_bar}) is not equivalent to a single geometry whose atomic positions are averaged over all conformers. The reason is that $d_{ij}$ is a nonlinear function of the positions $\mathbf{r}_i$ and $\mathbf{r}_j$, since $d_{ij} = \sqrt{(x_i-x_j)^2 + (y_i-y_j)^2 + (z_i-z_j)^2}$. Hence the average distance does not equal the distance between average positions. Equation (\ref{eq:d_bar}) then contains information of the position distribution up to at least its second order moment. Note that Eq. (\ref{eq:d_bar}) is invariant to translations or rotations of any one conformer, since each $d_{ij}^{(n)}$ is itself invariant.

\subsection{WHIM model}
To connect with recent 4D approaches in the drug discovery literature, we have also followed Ref. \cite{ash2017characterizing} and used the mean and standard deviation of conformer WHIM vectors for training. For each species we computed the WHIM vector of each conformer using RDKit, and computed the mean and standard deviation using the conformers' statistical weights.  These features were then combined with learnable fingerprints generated by ChemProp. As is the default for external features supplied to ChemProp, each of the WHIM features was scaled to have zero mean and variance. Our approach differs somewhat from Ref. \cite{ash2017characterizing} in that our conformers are generated in vaccuum rather than sampled from ligand-receptor MD, and also that we concatenate the WHIM features with those generated from message-passing.

\subsection{Baseline models}
We compared the 3D models to the 2D-based ChemProp and random forest models, with the random forest trained on Morgan fingerprints. We experimented with using atom-pair fingerprints \cite{carhart1985atom} for the random forest training, but did not find any difference in performance. We also experimented with a variant of the Compass approach to finding bioactive conformations \cite{jain1994compass}, training feed-forward networks on WHIM vectors instead of on ligand-query volume differences. We then used the predicted bioactive conformers as input to single conformer models. However, we discarded this approach because, for some targets, the model performance decreased as the predicted poses were updated. This indicated that the quality of the poses became worse, and hence that the iterative procedure was non-convergent.

\subsection{Obtaining conformers}
\label{subsec:conf}
To train a model on 3D ensemble information, one must first reduce the infinite set of 3D structures to a finite set of conformers, the set of thermally accessible structures at local minima on the potential energy surface. However, generating accurate conformers is challenging. A variety of exhaustive, stochastic, and Bayesian methods have been developed to generate conformers \cite{balloon_1, balloon_2, confab, frog2, moe, omega, rdkit, chan2019bayesian}. The stochastic conformer generation methods used in cheminformatics packages \cite{rdkit} are not exhaustive and may miss low-energy structures, and exhaustive sampling has prohibitive exponential scaling with the number of rotatable bonds. Further, the classical force fields used are generally not accurate enough to assign conformer orderings based on energy \cite{Kanal2018}.

Significant progress in conformer generation has been made with the CREST program \cite{crest}, which uses semi-empirical quantum chemistry and advanced sampling methods to generate reliable geometries. Conformer accuracy is crucial for all models that use 3D geometries. For example, in the QSAR Compass approach one considers only conformers with energy below 2.0 kcal/mol as candidates for the bioactive conformation \cite{jain1994compass}. However, because classical force fields cannot reliably assign energies in this narrow window \cite{Kanal2018}, it is easy to include high-energy conformers and miss low-energy ones. The CREST conformer orderings are more reliable because the energies are generated with semi-empirical DFT, which is a good approximation to high-accuracy \textit{ab initio} methods \cite{axelrod2020geom}. Further, stochastic methods can miss conformers, and MD simulations can become stuck in local minima even with long simulations. Indeed, the latter was suggested as one reason for the poor performance of the per-MD-frame WHIM vectors used in Ref. \cite{kyaw2020benchmarking}. These issues are avoided in CREST through the use of multiple rounds of metadynamics, high-temperature MD and genetic methods \cite{crest}. 

We recently used CREST to create the GEOM dataset  \cite{axelrod2020geom, geom_dataverse, geom_git}, which contains conformers annotated by energy for over 300,000 drug-like molecules with experimental binding data and 130,000 combinatorially generated small molecules \cite{qm9}. The 300,000 drug-like molecules contain ligands with \textit{in-vitro} data for the inhibition of SARS-CoV 3CL protease, SARS-CoV PL protease, SARS-CoV-2 3CL protease, SARS-CoV-2, \textit{E. Coli} and \textit{Pseudomonas aeruginosa}  \cite{ellinger, touret2020vitro, diamond_light, aid_binarized_sars_source, aid_485353, aid_652038, ecoli_1, ecoli_2, pseudomonas}. We  use this dataset to train the 3D conformer-based models introduced in this work.

\subsection{Training}

We trained different models on experimental data for the specific inhibition of the SARS-CoV-2 3CL protease (``CoV-2 CL'') \cite{diamond_light}, the general inhibition of SARS-CoV-2, measured \textit{in vitro} in human cells \cite{ellinger, touret2020vitro}, and the specific inhibition of the SARS-CoV 3CL protease (``CoV CL''). The CoV 3CL protease is found in the virus causing SARS, and has 96\% sequence similarity to CoV-2 CL \cite{cov_3cl_96_pct}. Because there is 1000 times more data for CoV 3CL than for CoV-2 3CL (Table \ref{tab:datasets}), and because the sequences are so similar, we also explored using transfer learning (TL) from CoV 3CL to CoV-2 3CL. Performance was measured using the receiver operating characteristic area under the curve (ROC-AUC), the precision-recall area under the curve (PRC-AUC), and the ROC enrichment (ROCE) \cite{jain2008recommendations} at 0.5\%, 1\%, 2\% and 2\%. Further discussion of these metrics can be found in the Supplementary Material (SM).

We used the conformers from the GEOM dataset as inputs to different 3D/4D models (SchNetFeatures, ChemProp3D, ChemProp3D-NDU). The training was performed with one conformer, with up to 200 conformers, and with a single effective conformer. We also trained a  ChemProp + WHIM model, which contains non-learnable conformer ensemble information. For comparison we also trained models with only graph information (ChemProp, and random forest over circular fingerprints). These networks were first trained on the CoV 3CL task and the two CoV-2 tasks. The CoV 3CL models were then used to generate fixed features for molecules in the CoV-2 3CL dataset for TL. Details of the data preprocessing, hyperparameter optimization, and uncertainty quantification of the model scores can be found in the SM.

\begin{table}[]
\centering
\scalebox{0.8}{
    \begin{tabular}{lcccc}
         \toprule
         & CoV 3CL & CoV-2 3CL  & CoV-2 \\ 
         \midrule 
         Train & 275 (167,255) & 50 (485) & 53 (3,294) \\
         Validation & 81 (55,750) & 15 (157) & 17 (1,096) \\
         Test & 70 (55,756) & 11 (162) & 22 (1,086) \\
         \midrule
         Total & 426 (278,758) & 76 (804) & 92 (5,476) \\
         \bottomrule
    \end{tabular}
    }
    \caption{Number of hits (total number of species in parentheses) for each split of each dataset. }
    \label{tab:datasets}
\end{table}

\section{Results}

ROCE scores of various models are shown in Fig. \ref{fig:roce}, and ROC and PRC scores are given in Table \ref{tab:non_tl_results}. For visualization purposes we have also included thin horizontal bars for each ROCE. The thin bars of identical ROCE scores are shifted down to avoid overlap. Bars stacked vertically with no spaces between them all have the value of the highest bar in the stack. For example, thin bars at negative $y$-values are always part of a stack for which the highest value is $\geq 0.$

\subsection{CoV-2 3CL}
\begin{figure}
    \centering
    \begin{tikzpicture}
    \node[inner sep=0pt] at (-3,0.8)
    {\includegraphics[height=0.35\textwidth,trim={11 0 0 0},clip]{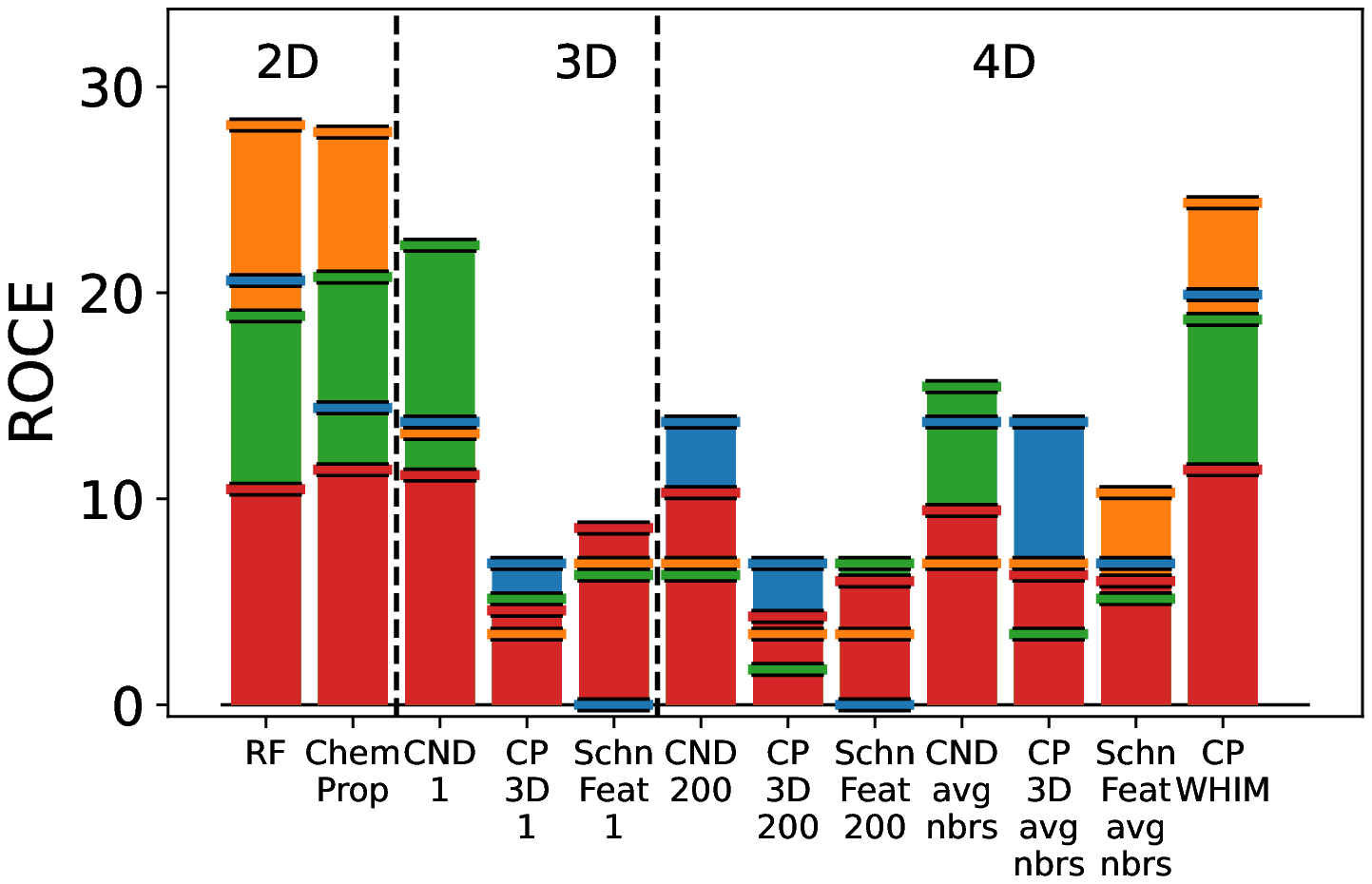}};
    \node[inner sep=0pt] at (4.0,0.8)
    {\includegraphics[height=0.35\textwidth,trim={30 0 0 0},clip]{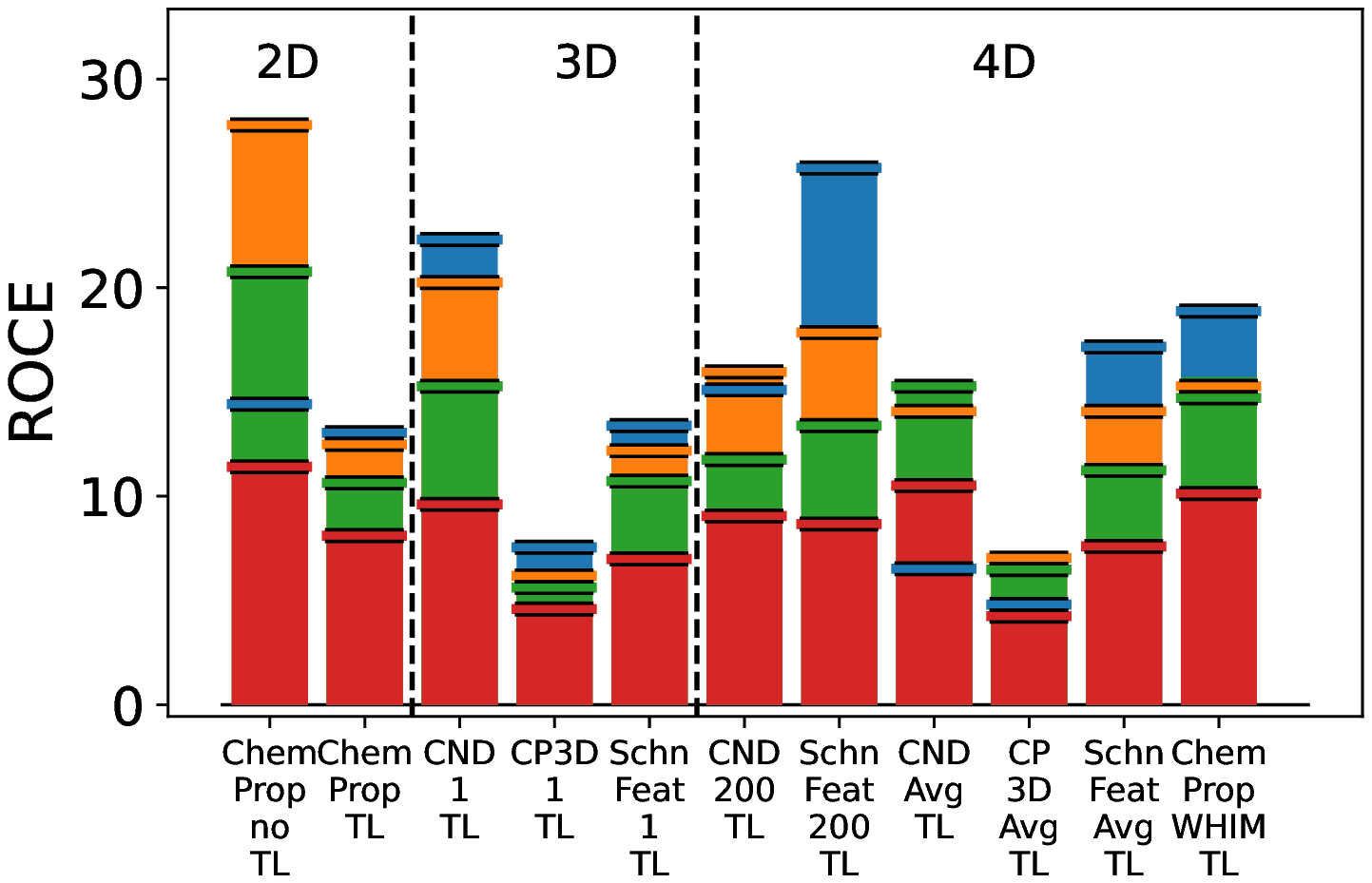}};
    
    \node[inner sep=0pt] at (0.55,-2.2)
    {\includegraphics[width=0.25\textwidth,trim={255 230 20 15},clip]{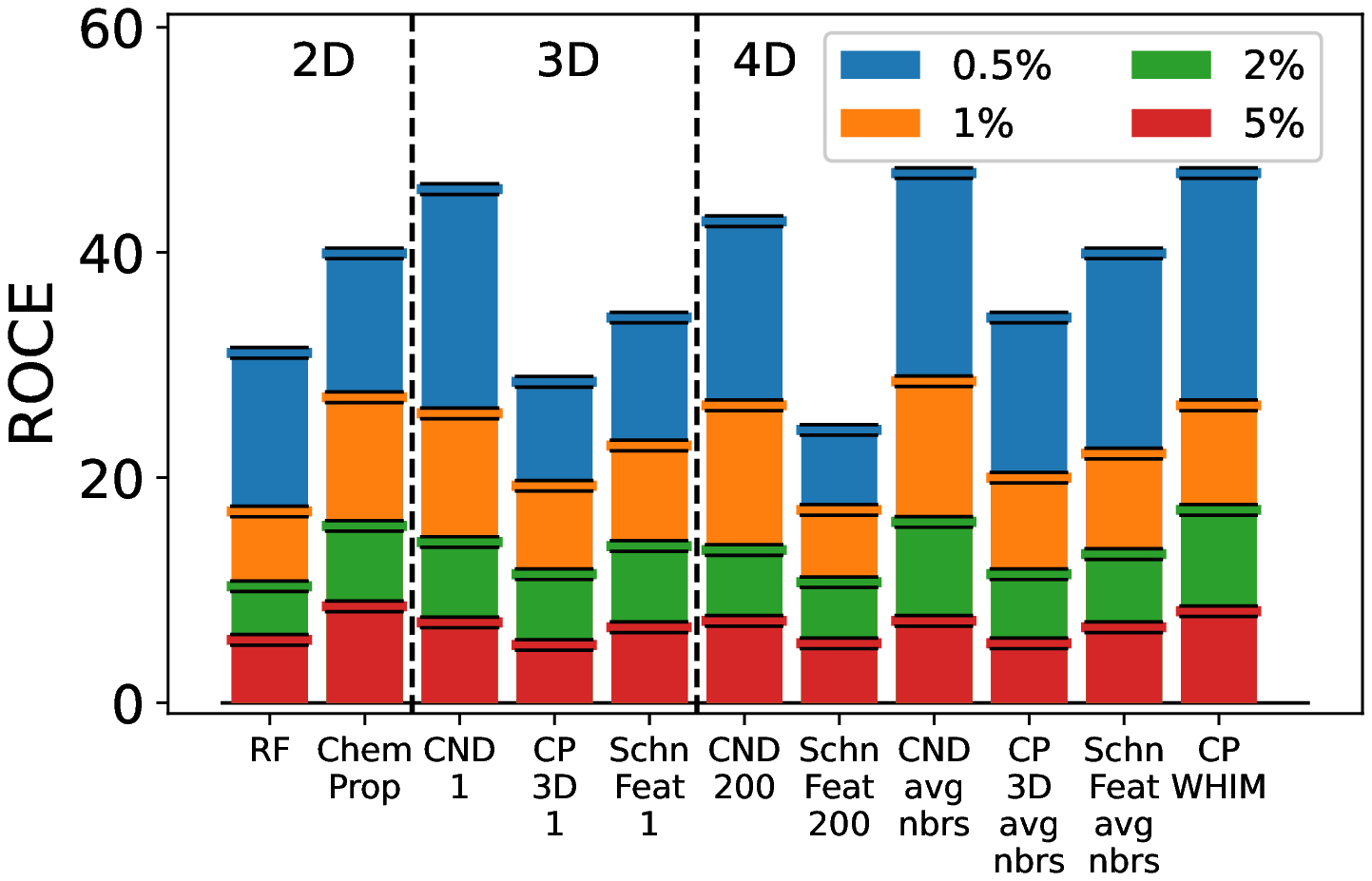}};

    \node[inner sep=0pt] at (-3,-5.4)
    {\includegraphics[height=0.35\textwidth,trim={11 0 0 0},clip]{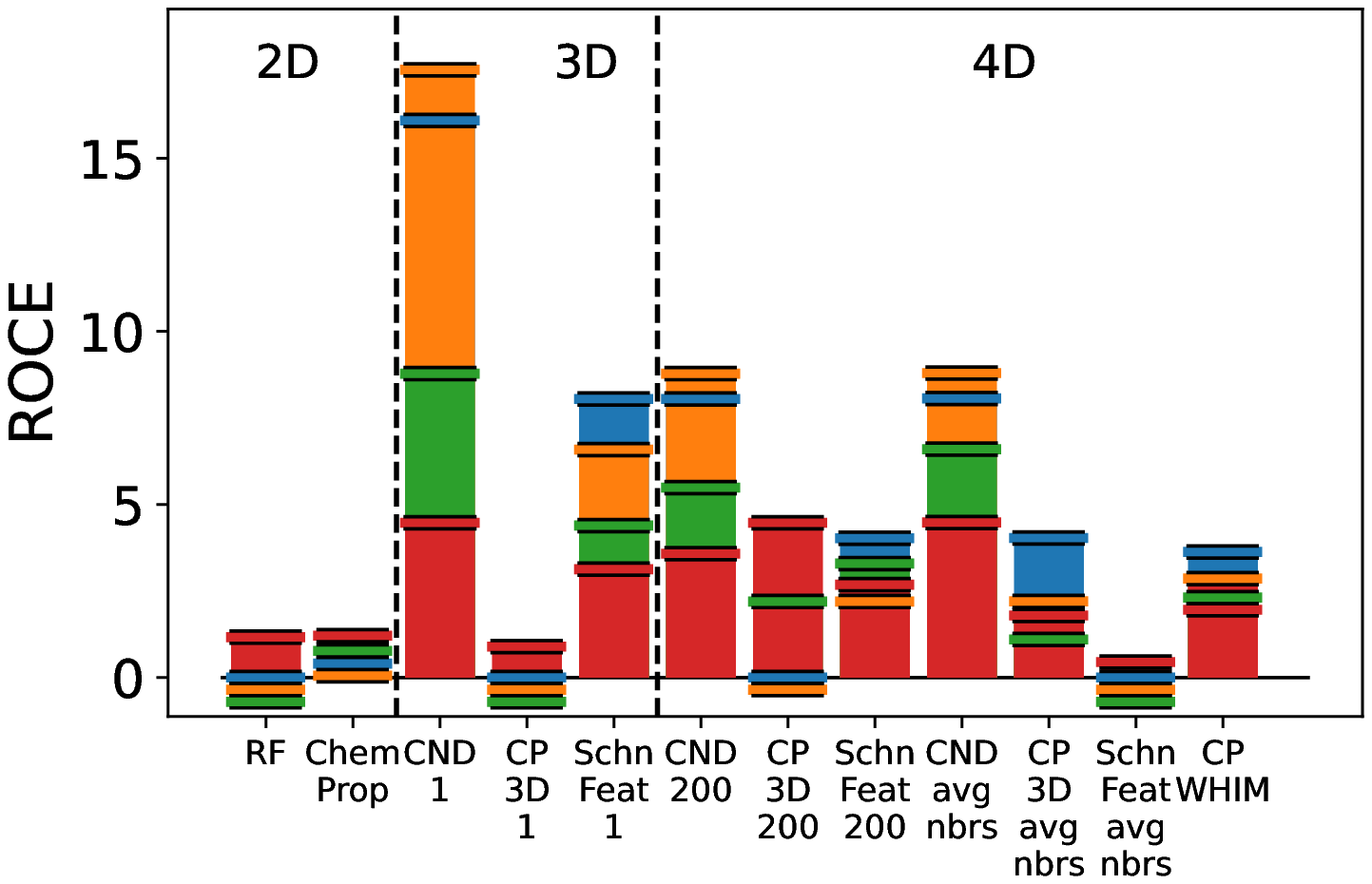}};
    \node[inner sep=0pt] at (4.0,-5.4)
    {\includegraphics[height=0.35\textwidth,trim={30 0 0 0},clip]{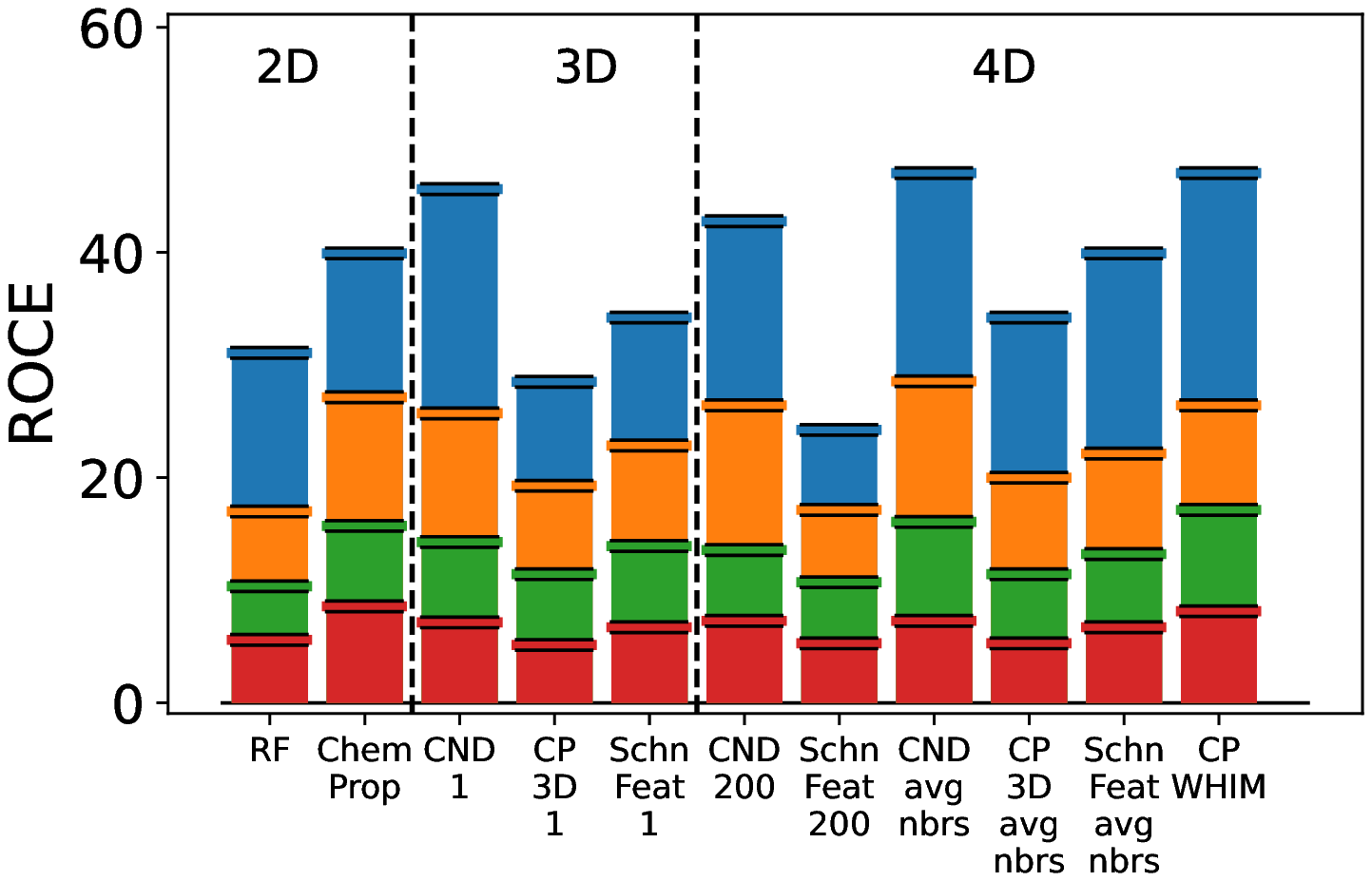}};

    \node[inner sep=0pt] at (-0.3, 2.6){(a)};
    \node[inner sep=0pt] at (6.55, 2.6){(b)};
    \node[inner sep=0pt] at (-0.3, -3.6){(c)};
    \node[inner sep=0pt] at (6.55, -3.6){(d)};
    
    \end{tikzpicture}
    \caption{ROCE scores of various models for different targets. The color of each bar indicates the false positive rate according to the central legend. (a) CoV-2 3CL protease, (b) CoV-2 3CL protease with transfer-learned models, (c) CoV-2, and (d) CoV 3CL protease.}
    
    \label{fig:roce}

\end{figure}

We begin by analyzing performance on CoV-2 3CL, the task with the smallest amount of data, and then proceed to datasets of increasing size (CoV-2 followed by CoV 3CL). Figure \ref{fig:roce}(a) shows that the 2D models generally have the best ROCE scores, while the 3D and 4D models are generally no better than the 2D models. The only exception is CND with one conformer (``CND-1'') at 2\% FPR, where its ROCE of 22.3 is slightly higher than ChemProp's ROCE of 20.8. The ChemProp model with WHIM features averaged over all conformers (``CP WHIM'') performs similarly to ChemProp and random forest. All other 3D and 4D models are significantly outperformed by 2D models. 

We also see that using MPNNs in the 2D models is largely unnecessary, as a simple random forest has comparable performance to the ChemProp models. For example, the scores and rankings of random forest and ChemProp are, respectively, 20.6/first and 14.4/third; 28.1/first and 27.8/second; 18.9/third and 20.8/second; and 10.5/fourth and 11.4/first. The random forest is the best model at 0.5\% and 1\% FPR, and is only slightly outperformed by ChemProp at higher FPR.

The ROCE scores are consistent with the ROC and PRC scores in Table \ref{tab:non_tl_results}. Here we see that ChemProp, random forest, CND-1 and CP WHIM all have high ROC and PRC scores. In fact, the CND-1 model has the highest ROC score while the CP WHIM model has the highest PRC. The other CND models also perform well, with CND-200 and CND avg achieving better ROC scores than ChemProp, and reasonably high PRC scores. 

This task has the smallest amount of data (485 training species), which may explain the results: with so little data, it is difficult for the neural networks to learn meaningful representations. The learned representations are likely no more powerful than the Morgan fingerprints used in random forest; this is often the case in low-data regimes \cite{pappu2020making}. This is especially true of the 3D models with distance updates, as they haven an abundance of input information with few labels to guide the representation learning. This may make the 3D models especially prone to overfitting. 

\renewcommand{\arraystretch}{1.3}

\begin{table}
\centering
\scalebox{0.8}{
\begin{tabu}{l|c|ccc|cc}

     \toprule
      \multicolumn{1}{c}{} & \multicolumn{1}{c}{} & \textbf{CoV-2 3CL} & \textbf{CoV-2} & \multicolumn{1}{c}{\textbf{CoV 3CL} }    \\
     \midrule
     & ChemProp  & 0.89$_{(1)}$ & 0.62$_{(4)}$ & 0.754$_{(7)}$ & \textbf{2D} \\ 
     & Random Forest & 0.92$_{(1)}$  & 0.61$_{(1)}$ & 0.69$_{(1)}$  \\ 

     \cmidrule{2-6} & SchNetFeatures (1 conf)  & 0.907$_{(5)}$ & \textbf{0.676}$_{(4)}$  & 0.718$_{(7)}$ \\ 
     & ChemProp3D (1 conf)  & 0.78$_{(5)}$ & 0.50$_{(1)}$ & 0.65$_{(2)}$ & \textbf{3D}   \\ 
      & CP3D-NDU (1 conf) & \textbf{0.929}$_{(3)}$ &  0.67$_{(0)}$  & 0.72$_{(3)}$ \\ 
      
     \cmidrule{2-6} & SchNetFeatures (200 confs) & 0.86$_{(2)}$ &0.63$_{(2)}$  & 0.72$_{(1)}$ \\ 
      \textbf{ROC} & ChemProp3D (200 confs)  &   0.66$_{(7)}$& 0.53$_{(0)}$  & --- \\ 
      & CP3D-NDU (200 confs) & 0.901$_{(4)}$ & 0.663$_{(8)}$ & 0.759$_{(2)}$  \\
      
      \tabucline [1pt on 0.5pt off 2pt]{2-5} 
      
      & SchNetFeatures (avg nbrs)  & 0.84$_{(3)}$ & 0.61$_{(1)}$ & 0.74$_{(2)}$ & \textbf{4D} \\ 
      & ChemProp3D (avg nbrs)  & 0.73$_{(0)}$ & 0.56$_{(2)}$ & 0.70$_{(3)}$ &  \\ 
      & CP3D-NDU (avg nbrs) & 0.916$_{(5)}$ & 0.647$_{(3)}$  &  0.71$_{(0)}$ &  \\
      
      \tabucline [1pt on 0.5pt off 2pt]{2-5} 
      & ChemProp + WHIM  & 0.89$_{(2)}$ & 0.66$_{(5)}$ & \textbf{0.77}$_{(7)}$ & \\
      
     \midrule\midrule
     & ChemProp  & 0.56$_{(6)}$ & 0.028$_{(3)}$  & 0.05$_{(1)}$  & \textbf{2D} \\
     & Random Forest   & 0.55$_{(3)}$ & 0.028$_{(2)}$ & \textbf{0.079}$_{(2)}$ \\

     \cmidrule{2-6} & SchNetFeatures (1 conf) & 0.333$_{(0)}$ & 0.051$_{(3)}$ & 0.021$_{(6)}$   \\ 
     & ChemProp3D (1 conf) & 0.33$_{(6)}$ & 0.021$_{(0)}$ &  0.04$_{(3)}$  & \textbf{3D} \\ 
     & CP3D-NDU (1 conf)  & 0.533$_{(1)}$ & \textbf{0.124}$_{(0)}$ & 0.07$_{(2)}$ \\ 
     
     \cmidrule{2-6}  & SchNetFeatures (200 confs) & 0.26$_{(5)}$  & 0.037$_{(3)}$ & 0.03$_{(2)}$   \\
      \textbf{PRC} & ChemProp3D (200 confs) & 0.20$_{(1)}$ & 0.032$_{(0)}$ &  ---  \\ 
      & CP3D-NDU (200 confs)  & 0.413$_{(3)}$ & 0.06$_{(1)}$  & 0.04$_{(3)}$ \\ 

      \tabucline [1pt on 0.5pt off 2pt]{2-5} 
      
      & SchNetFeatures (avg nbrs)   & 0.29$_{(6)}$ & 0.027$_{(0)}$ & 0.07$_{(3)}$ & \textbf{4D} \\ 
      & ChemProp3D (avg nbrs)  & 0.31$_{(0)}$ &  0.10$_{(6)}$  & 0.07$_{(1)}$ &  \\ 
      & CP3D-NDU (avg nbrs)  & 0.467$_{(5)}$ & 0.058$_{(4)}$ & 0.063$_{(1)}$  &  \\
      
      \tabucline [1pt on 0.5pt off 2pt]{2-5} 
      & ChemProp + WHIM & \textbf{0.57}$_{(6)}$ & 0.04$_{(1)}$ & 0.07$_{(2)}$   &  \\
      
     \bottomrule
\end{tabu}
}
\caption{Performance of various models on the three classification tasks, evaluated with the ROC and PRC scores. Uncertainty on the last digit is denoted by a subscript, and is given here by the standard deviation of test scores. Uncertainties less than $10^{-3}$ are recorded as zero. Due to computational constraints we did not train ChemProp3D (200 confs) on the CoV 3CL task.}
\label{tab:non_tl_results}
\end{table}



\subsection{CoV-2}

The results are quite different for the CoV-2 task. As seen in Fig. \ref{fig:roce}(c), the ChemProp and random forest models are outperformed by a number of 3D models. The CND-1 model in particular has the highest ROCE scores at 0.5\%, 1\%, and 2\% FPR, and is essentially tied with CND avg and CP3D-200 for the highest score at 5\% FPR. The CND-1 model outperforms ChemProp, the best 2D model, by a significant margin. At 0.5\% FPR, CND-1 has an ROCE of 16.1, nearly 40 times higher than ChemProp's ROCE of 0.4. At lower FPR values the improvement is smaller but still quite significant, with CND-1 outperforming ChemProp by factors of 26, 11, and 4 at FPRs of 1\%, 2\%, and 5\%, respectively.

Other 3D and 4D models also perform better than ChemProp, though none achieves the success of CND-1. For example, SchNetFeat-1, CND-200, and CND avg all achieve high ROCE scores, improving on the ChemProp scores by factors of 4-20 at various FPR values. WHIM fingerprints also improve the performance of ChemProp, increasing ROCE scores by factors between 3 and 9. This should be contrasted with the CoV-2 3CL task, in which adding WHIM fingerprints did not improve performance. We also note that the random forest model performs quite poorly, achieving zero ROCE for 0.5\% and 1\% FPR, and an ROCE of 0.22 at 2\% FPR (three times lower than ChemProp). Evidently message-passing is far more helpful here than in the CoV-2 3CL task.

These trends are reflected in the PRC scores, and to a lesser extent in the ROC scores. Table \ref{tab:non_tl_results} shows that SchNetFeat-1, CND-1, CND-200, and CND-avg all have some of the highest ROC scores. However, the same is true of CP WHIM, which has a rather low ROCE score. Moreover, ChemProp and random forest, which have the lowest ROCE scores, have ROC values only slightly below those of the best models (approximately 8\%). Clearly the ROC score is insufficient for evaluating the models. By contrast, the PRC scores show that ChemProp and Random Forest are significantly outperformed by SchNetFeat-1, CND avg and CP WHIM, which in turn are significantly outperformed by CND-1. Indeed, the CND-1 score is four times that of ChemProp and double that of SchNetFeat-1, CND avg and CP WHIM. The CP3D avg model also has a very high PRC, though interestingly its ROC is rather low. We also note that while CND-1 performance is quite impressive, the CND-1 model with the best validation PRC also had the best validation ROC, so the uncertainty for each of its scores is zero (see SM). It is therefore possible that the extremely high PRC and ROCE scores are somewhat lucky. However, even if we ignored the CND-1 model, the next best 3D models would still be significant improvements over the 2D models. 

For this task it is clear that 3D information can significantly improve performance. Two factors distinguish this task from the CoV-2 3CL problem. First, there is substantially more data (see Table \ref{tab:datasets}). Second, the hits may have completely different mechanisms of action from each other, as they inhibit the \textit{in-vitro} growth of SARS-CoV-2 in human cells, while the CoV-2 CL data is for ligands that specifically inhibit the CoV-2 3CL protease. Indeed, the ROCE scores are far lower than the best ROCE scores in the CoV-2 3CL task, indicating that this problem is more difficult. Below we show that MPNN-based 3D and 4D models also perform well for the CoV-3CL classification, but that they are only slightly better than 2D models. Therefore, it is possible that 3D information is more important when the mechanisms of action are different, and/or when there is an intermediate amount of training data (not too much or too little). Further investigation is required to test these hypotheses.

Another important observation is that the 4D models are no better than the 3D models. For example, the 4D models with the best ROCE scores are CND-200 and CND avg, but both are outperfomed by CND-1 at FPR values below 5\% and matched at 5\%. Similarly, SchnetFeat-200 is outperformed by SchNetFeat-1 at every FPR. The exception to this trend is CP3D avg, which is more accurate than CP3D-1. However, it is still far worse than the other 3D models. These trends can also be seen in the ROC and PRC scores. 

Since all information contained in a single conformer representation is also available with 200 conformers, it seems that the model training is hampered by the extra information. This is particularly intriguing because the attention mechanism learns meaningful information about the conformers (see below), focusing on geometries that are similar to those of other hits. The model may be overfitting to the extra information, despite the hyperparameter optimization to determine reasonable dropout rates.

\subsection{CoV 3CL}

The final target is CoV 3CL. It contains by far the largest training set, and, like CoV-2 3CL, contains experimental data for ligands that bind to a specific protein. According to the ROCE scores in Fig. \ref{fig:roce}(d), the models that perform best on average are CND avg, CP WHIM, CND-1, CND-200, and ChemProp. The CND avg is best at FPRs of 0.5\% and 1\% (ROCE=47.0 and 28.6), CP WHIM is best at 2\% (ROCE=17.1), and ChemProp is best at 5\% (ROCE=8.6). 

The 3D/4D models are generally better than the 2D models, but only by a small margin. For example, the ratios of top 3D/4D score to top 2D score are 1.18, 1.05, 1.09, and 0.95 at FPRs of 0.5\%, 1\%, 2\% and 5\%, respectively. Interestingly, this is the only task in which 4D models consistently outperform 3D models, but the margin is again rather small. Specifically, the ratios of top 4D score to top 3D score are 1.84, 1.11 1.20, and 1.14 at FPRs of 0.5\%, 1\%, 2\% and 5\%, respectively. 

We also see that the random forest is outperformed by MPNN models, though to a lesser extent than in the CoV-2 task. The superiority of MPNNs is to be expected for this task because the dataset is quite large. This provides ample information from which to learn an ideal molecular representation. The size of the dataset also likely explains the fact that the ROCE scores are higher than in any other task.

The ROCE trends are only somewhat similar to the PRC and ROC trends. The random forest model has the highest PRC score, and the CND-1, SchNetFeat avg, CP3D avg, CND avg and CP WHIM models are close behind. The ChemProp PRC is noticeably lower than each of these models, while CND-200 score is lower still. While the CND-1, CND avg and CP WHIM models also have high ROCEs, the CP3D avg and random forest models do not. In fact, according to the ROCE metric, the random forest is one of the worst models. Moreover, ChemProp and CND-200 have high ROCE scores but low PRCs. These results again emphasize the importance of going beyond the ROC and PRC scores when comparing models.

\subsection{Transfer learning}
\begin{table}[t]
    \centering
	\scalebox{0.8}{

	\begin{tabu}{lcc|cc}
            \toprule
            & CoV 3CL model & CoV-2 3CL message & ROC & PRC \\ 
            & for fingerprinting & passing (Y/N) && \\
            \midrule
            & ChemProp (2D) & Y & 0.78$_{(3)}$ & 0.36$_{(8)}$  \\
           & ChemProp (2D) & N & 0.71$_{(6)}$ & 0.35$_{(7)}$ \\
            \midrule
            
            & SchNetFeatures (1 conf) & Y & 0.76$_{(8)}$ & 0.3$_{(1)}$ \\
            & SchNetFeatures (1 conf ) & N & 0.77$_{(7)}$ & 0.3$_{(2)}$ \\

            \tabucline [1pt on 0.5pt off 2pt]{1-5} 
            & ChemProp3D (1 conf) & Y & 0.75$_{(5)}$ & $0.2_{(2)}$ \\
            & ChemProp3D (1 conf) & N & 0.69$_{(6)}$ & 0.2$_{(1)}$ \\

             \tabucline [1pt on 0.5pt off 2pt]{1-5}
            & CP3D-NDU (1 conf) & Y & 0.85$_{(3)}$ & 0.43$_{(9)}$  \\
            & CP3D-NDU (1 conf) & N & 0.81$_{(4)}$ & 0.47$_{(9)}$ \\

            \midrule
            
            & SchNetFeatures (200 confs) & Y  & 0.86$_{(6)}$ & 0.42$_{(9)}$ \\
            & SchNetFeatures (200 confs) & N & 0.87$_{(5)}$ & \textbf{0.50}$_{(9)}$ \\
            
            \tabucline [1pt on 0.5pt off 2pt]{1-5} 
            & CP3D-NDU (200 confs) & Y & \textbf{0.88}$_{(5)}$ & 0.47$_{(9)}$ \\
            & CP3D-NDU (200 confs) & N & 0.87$_{(5)}$ & 0.4$_{(1)}$ \ \\
            \tabucline [1pt on 0.5pt off 2pt]{1-5}

            & SchNetFeatures (avg nbrs) & Y & 0.79$_{(6)}$ & 0.3$_{(1)}$ \\
            & SchNetFeatures (avg nbrs) & N & 0.83$_{(6)}$ & 0.4$_{(1)}$\\

            \tabucline [1pt on 0.5pt off 2pt]{1-5} 
            & ChemProp3D (avg nbrs) & Y  & 0.6$_{(2)}$ & 0.2$_{(1)}$ \\
            & ChemProp3D (avg nbrs) & N & 0.74$_{(6)}$ & 0.29$_{(9)}$ \\

             \tabucline [1pt on 0.5pt off 2pt]{1-5}
            & CP3D-NDU (avg nbrs) & Y & \textbf{0.88}$_{(2)}$ & 0.48$_{(5)}$ \\
            & CP3D-NDU (avg nbrs) & N & 0.87$_{(3)}$ & 0.41$_{(5)}$   \\
            
             \tabucline [1pt on 0.5pt off 2pt]{1-5}
            & ChemProp + WHIM & Y  & 0.77$_{(3)}$ & 0.42$_{(4)}$ \\
            & ChemProp + WHIM & N  & 0.81$_{(2)}$ & 0.44$_{(3)}$ \\

            \bottomrule
            
        \end{tabu}
        
    }
    \caption{Transfer learning results from CoV 3CL to CoV-2 3CL for different models, with and without additional message passing from ChemProp.}
    \label{tab:cov_2_tranf}
\end{table}

We also analyzed how TL could improve CoV-2 3CL scores. To do so we used the pre-trained CoV 3CL models to generate fingerprints, and used the fingerprints as input to a standard feed-forward neural network. We did this with and without additional message passing from ChemProp. When additional message passing was used, the fixed fingerprints were concatenated with the learned ChemProp fingerprints. For each architecture we used two models, one with the best CoV 3CL AUC score and one with the best CoV 3CL PRC score. Each of the two pre-trained models was then used to produce molecular fingerprints, which were then used to train an ensemble of ten different CoV-2 3CL models. The final scores are thus an average over 20 different models. Hyperparameters were again optimized separately for each model. 

Transfer learning ROC and PRC score are shown in Table \ref{tab:cov_2_tranf}, and can be compared with the ChemProp and random forest baselines in Table \ref{tab:non_tl_results}. The ROCE scores are shown in Fig. \ref{fig:roce}(b), and correspond to averages over 40 models (20 with additional message-passing and 20 without). The ROC and PRC scores show no evidence that TL improves performance, as the best TL scores are lower than those of ChemProp and Random Forest without pre-training. However, the ROCE scores show that TL helps at low FPR. In particular, SchNetFeat-200 TL at 0.5\% FPR has an ROCE of 25.7, which is 1.25 times greater than that of random forest (ROCE=20.6), and 1.78 times greater than that of ChemProp (ROCE=14.4). CND-1 TL also shows strong performance, as its ROCE of 22.3 is 1.08 times larger than random forest and 1.54 larger than ChemProp. The strong performance of these models is consistent with their high ROC and PRC scores, though the latter never exceed those of non-TL models. At FPR values above 0.5\% there is no advantage to using transfer learning, as the highest scores are always those of non-TL models.

It is interesting that 3D/4D TL models perform better than the 2D TL model. This is reflected in the ROCE, ROC, and PRC scores. It seems that the representations learned by the 3D models are more easily generalized to new targets than the 2D representations. It is also noteworthy that SchNetFeat-200, a 4D model, provides the best TL results. However, by all metrics SchNetFeat-200 is one of the worst models at predicting CoV 3CL inhibition, while CND-1 is one of the best. It is therefore difficult to explain why they provide some of the best TL results at low FPR.

The ROCE improvements at 0.5\% FPR are certainly interesting, but they are not enough to justify pre-training a 3D or 4D model with hundreds of thousands of species. The non-TL ChemProp model performs better at all other FPR values, and a random forest with Morgan fingerprints has competitive accuracy. Further work is needed to evaluate the usefulness of TL in drug discovery.

\section{Discussion}

\subsection{Accuracy}

Adding 3D information can improve prediction accuracy when using the right model. Of all the models introduced, the new CND model shows the strongest and most consistent performance, either matching or beating 2D baselines in most tasks. CND (1-C) in particular shows the most consistent performance of all models, even on the CoV-2 3CL task with little data. The CND model is similar to PotentialNet \cite{feinberg2018potentialnet}, in that both contain a separation between distance updates and graph updates. PotentialNet uses two sets of gated graph neural networks (GGNNs) \cite{li2015gated}, the first of which acts on the 2D graph alone. The output of this network is a set of node features $h_v^{T}$ that have been updated $T$ times. The second GGNN updates the node features $h_v^{T}$ through convolutions over the 2D graph and the 3D structure. Using the $h_v^{T}$ as input to the second network means that graph information is already present in the nodes once 3D information is added. Hence the second network can fine-tune the graph-based features using 3D information, rather than generating all features at once. The CND approach also separates the aggregation of graph and distance information (see Methods above). While it does not include distance-based convolutions, its use of edge states means that distance information is included for all pairs within $r_{\mathrm{cut}}$ of each other. This is roughly equivalent to a single 3D convolution in a node-based model.

While the CND model matches or improves upon 2D baselines, many 3D models do not improve performance, and some can actually hurt it. For example, ChemProp3D is outperformed by 2D models in every task, while SchNetFeatures improves results in CoV-2 but hurts them in CoV 3CL. It is also clear that adding extra conformers hurts performance in most tasks (CoV-2 3CL and CoV-2), and only leads to small improvements in others (CoV 3CL). Given the enormous computational cost of training the 200-C models (see below), it is a far better use of resources to train single conformer models. Transfer learning can somewhat improve accuracy, but the effect is small and absent in most models. Much of the improvement can be matched with a simple 2D baseline like random forest.

The results of MPNN-based 3D models and 3D models with non-learnable features are also mixed. For CoV-2 3CL it is best to use a 2D model or a 3D model with non-learnable WHIM features. For CoV 3CL it is best to use a 3D or 4D model with learnable features, but using WHIM features gives comparable performance. By contrast, for CoV-2 it is far better to use MPNN-based 3D models. This is another example of a trend that is different for CoV-2 than for the other targets. It therefore reinforces the need for further investigation into tasks without a well-defined protein target.

\subsection{Conformer importance} 

Using 4D models leads to minor improvement over 3D models for CoV 3CL, but hurts performance for other targets. Moreover, at each FPR value the best 4D model is CND avg or CP WHIM, neither of which uses attention to determine the most important conformation. It is therefore important to understand how conformers are pooled to see if the models are learning meaningful information from them. To this end we analyzed the similarity of high-attention conformers in different species. For each comparison we randomly selected two species, and for each species we selected the conformer with the highest attention weight among all attention heads. We then computed the fingerprints of the two conformers using the E3FP method \cite{e3fp}, which is an extension of the extended connectivity fingerprint to 3D structures. The cosine similarity metric was then applied to the fingerprints. This was repeated for 5,000 random pairs in the test set to obtain the average similarity. We applied this analysis separately to pairs of species that were both hits, and to pairs of species that contained one hit and one miss. We also repeated the analysis with a random selection of conformers for comparison.

Results of this analysis are shown in Table \ref{tab:att_sim}.\footnote{In the table we have reported the uncertainty as the standard error of the mean (SEM). The standard deviation would indicate the expected range of scores for any \textit{single} pair of species, and thus the likelihood of the attention model beating the random model for any one pair. By contrast, the SEM measures our confidence that the \textit{average} difference between the models is statistically significant. The SEM values in Table \ref{tab:att_sim} suggest that the difference between the models is not random.} Random selection of conformers yields hits that are more similar to each other than to misses. Choosing conformers based on attention increases the similarity of hits to each other, while leaving the similarity of hits to misses virtually unchanged. The difference between hit/hit and hit/miss scores, denoted by $\Delta$, is on average 40\% larger for conformers chosen by attention than conformers chosen randomly. This means that it is easier to distinguish hits from misses using attention conformers than using random conformers. Interestingly, unlike the other models, CND (200-C) did not learn meaningful attention weights for the CoV 3CL task (it assigned equal weights to nearly every conformer), but had the strongest performance of any attention-based 4D model. This result, along with the strong performance of the attention-free CP WHIM and CND avg models, suggest that best results may be achieved without attention. However, the clear evidence that attention weights contain meaningful conformer information indicate that further investigation is required.


\begin{table}[t]
\centering
\scalebox{0.8}{
\begin{tabular}{c|c|c|c||c|c|c||c|c|c}
     \multicolumn{1}{c}{} & \multicolumn{1}{c}{} & \multicolumn{1}{c}{CoV} & &  \multicolumn{1}{c}{} & \multicolumn{1}{c}{CoV-2} & &  \multicolumn{1}{c}{} & \multicolumn{1}{c}{CoV-2 3CL} & \multicolumn{1}{c}{} \\
     \cmidrule{2-10}
      & hit/hit & hit/miss & $\Delta$ & hit/hit & hit/miss & $\Delta$ & hit/hit & hit/miss & $\Delta$  \\ 
     \midrule 
     Attention & 0.325$_{(1)}$ &  0.3071$_{(9)}$ & \textbf{0.018}$_{(1)}$ &  0.340$_{(2)}$ & 0.291$_{(1)}$ &\textbf{ 0.049}$_{(2)}$ & 0.378$_{(3)}$ & 0.276$_{(1)}$ & \textbf{0.102}$_{(3)}$ \\
     Random &  0.321$_{(1)}$ & 0.3054$_{(9)}$  & 0.015$_{(1)}$ & 0.323$_{(1)}$ & 0.293$_{(1)}$  &0.031$_{(2)}$ & 0.344$_{(2)}$ & 0.273$_{(1)}$ & 0.071$_{(2)}$ 
\end{tabular}}
\caption{Average cosine similarity between E3FP fingerprints of different species. A single conformer was selected for each species, either randomly or based on highest attention weight from a SchNetFeatures model with the lowest validation loss. ``hit/hit'' means similarity among hits, ``hit/miss'' means similarity between hits and misses, and $\Delta= \text{(hit/hit)} - \text{(hit/miss)}$. The highest $\Delta$ scores for each task are shown in bold. 5,000 comparisons were made for each category. Uncertainty on the last digit is denoted by a subscript and given by the standard error of the mean. }
\label{tab:att_sim}
\end{table}

    

    


\subsection{Computational cost}
The 4D attention-based models are quite computationally demanding. 3D models are generally more expensive than 2D models, as they incorporate non-bonded edges in addition to bonded ones. More importantly, conformer-based models are $n_{\mathrm{conf}}$ times more expensive than single conformer models, where $n_{\mathrm{conf}}$ is the average number of conformers. Here we used a maximum of 200 conformers per species, which corresponds to 76 on average. Hence our models are hundreds of times slower than 2D models, and our training on the CoV 3CL dataset took several days on 32 GPUs. 

It is not beneficial to use conformer models in their current state, as they are expensive and can hurt performance. However, given the sheer volume of extra information contained in conformers, it is helpful to think of other ways to use the ensemble. Let us consider how model complexity can be reduced, as this would reduce computational cost and could also improve performance. A simple method is to bin similar conformers together, so that the number of effective conformers is reduced. The statistical weight of each conformer would be the sum of its constituent weights. This would lead to models that act over far fewer conformers for each species, thereby reducing complexity and cost. One might also consider expanding upon the average-distance models introduced here, which come at the cost of single-conformer models but contain ensemble information. Performance could be improved by using pre-set weights that reflect binding affinity rather than statistical weight. These could perhaps be found with computational docking or with attention models acting over non-learnable 3D fingerprints, which are far less expensive. One might also use the docking results to simply select the best single conformer for each species, and train a 3D model on one conformer only for each species.

\section{Conclusion}
We have introduced a new method for predicting properties from conformer ensembles, and introduced several new 3D-based models. We achieved significant improvement over 2D models with a single conformer, but found that multiple conformers did not improve the predictions. With access to the GEOM dataset and the models introduced in this work, the community will be able to use our models and refine our conformer approach to improve virtual screening.

\section{Data and Software Availability}
All code and data is publicly available. The GEOM dataset can be accessed at \cite{geom_dataverse}, with instructions for loading the data in \cite{geom_git}. Code for training the models is available in the Neural Force Field repository at \cite{nff}. Training datasets, trained models, log files, and model details can be found at \cite{dataverse_models}.

\section{Acknowledgements}

The authors thank the XSEDE COVID-19 HPC Consortium, project CHE200039, for compute time. NASA Advanced Supercomputing (NAS) Division and LBNL National Energy Research Scientific Computing Center (NERSC), MIT Engaging cluster, Harvard Cannon cluster, and MIT Lincoln Lab Supercloud clusters are gratefully acknowledged for computational resources and support. The authors also thank Christopher E. Henze (NASA) and Shane Canon and Laurie Stephey (NERSC) for technical discussions and computational support, MIT AI Cures (https://www.aicures.mit.edu/) for molecular datasets and Wujie Wang, Daniel Schwalbe Koda, Shi Jun Ang (MIT DMSE) for scientific discussions and access to computer code. Financial support from DARPA (Award HR00111920025) and MIT-IBM Watson AI Lab is acknowledged.

\newpage
\title{Supplementary Material for Molecular machine learning with conformer ensembles}

\maketitle
\beginsupplement

\section{Saved datasets and models}
The code for creating datasets and training models is available in the Neural Force Field repository \cite{nff}, which contains a variety of 3D-based models \cite{schnet_1, schnet_2, Klicpera2020} to predict either molecular properties or atomic forces. The force field can handle periodic structures, generate coarse-grained forces \cite{ruza2020temperature, wang2019coarse}, and run molecular dynamics through the Atomic Simulation Environment \cite{ase}. The property predictor extends SchNet \cite{schnet_1, schnet_2} and ChemProp \cite{chemprop} to predict properties based on one or more conformer, and contains scripts with extensive documentation for dataset generation, training, hyperparameter optimization, and transfer learning.

The trained models and datasets themselves can be found at \cite{dataverse_models}. Both the model and dataset folders have sub-folders for each prediction task, with the model folder further split by network architecture. Each network folder contains log files with training and validation scores for each epoch, models saved at every epoch, the parameters used to create and train the model, fingerprints and attention weights generated for the test set, and test set predictions. There are also folders with ChemProp models trained from scratch, ChemProp models used for transfer learning, and parameter scores from hyperparameter optimization. Further details of the data layout can be found in the accompanying \texttt{README} file.
 
\section{Data pre-processing}

Because the conformers were generated with CREST, which allows for reactivity such as keto-enol tautomerism, some of the species had conformers with different molecular graphs from each other. 
A small number of conformers also had bond lengths greater than the cutoff distance. 
To avoid these issues we removed all such problematic species, which accounted for about 4\% of the total. 
This is why there are fewer species here than in the original sources, or in the associated datasets found in \cite{aicures}. The data used are available at \cite{dataverse_models}.

Due to computational constraints we limited the dataset to the 200 highest-probability conformers of molecules with 100 atoms or fewer. The probabilities were renormalized to account for the missing conformer probabilities. The constraint on the number of atoms excluded only 0.2\% of molecules. The conformer constraint was more restrictive, but still reasonable. 85\% of the molecules had 200 conformers or fewer. On average, a limit of 200 conformers recovered 94\% of the total number of conformers of a species.

\section{Training} 

In all cases we used a 60/20/20 train/validation/test split with species split by scaffold, as described in Ref. \cite{chemprop}. This approach groups molecules with the same scaffolds into the same split. This makes the classification task harder than a random split, as the model cannot simply identify a scaffold from the training set and apply it to the validation and test sets.

The training process was very computationally demanding. First, the dataset contained 30 million geometries. Second, the computational graph of each species contained up to 200 separate graphs for each fingerprint. Since the fingerprints of a species were pooled and used as input to the readout layer, none of the computational graphs could be broken until all fingerprints had been created and pooled. Third, the ChemProp3D models had to store and update edge features. For a molecule with $M$ atoms and an average of $m$ neighbors per atom, this means that $M \cdot m$ directed edge features had to be updated, whereas a node-based MPNN would only update $M$ node features. The average atom had $m=11$ neighbors. Moreover, the neighbors of each edge also had to be aggregated, which further increased computational cost.

To address these challenges we parallelized training over 32 Nvidia Volta V100 GPUs (32 GB of memory each), and for ChemProp3D models performed batching over conformers. Each batch consisted of 4-7 conformers for one species, which was the largest value that could fit in GPU memory. This value is rather small, as memory also had to be allocated for the computational graph of all conformers. Fingerprints were generated for each batch in succession. The fingerprints were pooled once all batches were finished. The pooled fingerprints were used as input for the readout layer, which generated a prediction. The prediction was used to update the loss gradient. The computational graph was erased, freeing memory to repeat the process. This process was repeated twice on each GPU, generating a loss gradient from 64 species. The gradient was then used for a training step.

3D models were trained with an initial learning rate of $10^{-4}$. The learning rate was reduced by a factor of two if the validation loss had not decreased in 10 epochs. The training continued until the learning rate reached  $10^{-6}$. The exception was CoV 3CL models, which were only trained for 25-50 epochs because of time constraints. The model with the best score on the validation set was evaluated on the test set. To account for the binding/non-binding class imbalance we used restrictive over-sampling. In this approach the under-represented class is sampled at an equal rate to the over-represented class. The average batch then contains half positive binders and half negative binders. One epoch samples $N$ molecules, where $N$ is the size of the dataset, but only about half of them are unique: the other half contains positives that are continually resampled. It therefore takes the model two epochs to see the entire dataset. In ChemProp an epoch contains all positives sampled once, with an equal number of negatives. It then takes the model $n_{\mathrm{neg}} /  n_{\mathrm{pos}}$ epochs to see all the data. Since ChemProp usually requires 30 epochs for convergence, we trained all ChemProp models for  $\approx 30 \ n_{\mathrm{neg}} /  n_{\mathrm{pos}}$ epochs, though the models usually converged far earlier.

\section{Hyperparameter optimization}

Hyperparameter optimization was performed for each model type and each individual task using Bayesian optimization \cite{hyperopt}. In all cases we scored models by their best validation scores rather than test scores, to avoid biasing the models toward molecules in the test set. Models were then trained on full datasets.

Hyperparameters for ChemProp models were optimized using the defaults in the repository \cite{chemprop_git}, which vary the dimension of the hidden state (300 to 2400), number of convolutions (2 to 6), dropout rate (0 to 0.4), and number of readout layers (1 to 3). For single conformer 3D models we optimized the dropout rate in the convolution layers (0 to 0.4) and readout layers (0 to 0.4), sampling the logarithm of the dropout rate uniformly. For multiple conformer species we also optimized the type of attention (linear or pair-wise), the dropout rate in the attention layer, and the number of attention heads. We used Bayesian hyperparameter optimization for each model \cite{hyperopt}, with 20 hyperparameter combinations for faster models and 5-10 for slower ones. For average neighbor models we used the hyperparameters that were optimized for single-conformer models with the same architecture. For example, for the CND average model we used the hyperparameters optimized for the CND (1-C) model.

To choose the best model we evaluated the score on the validation set (not the test set). In most cases the PRC-AUC metric was used, for the following reason. Both the PRC and ROC involve recall (or true positive rate), the proportion of total hits correctly identified by the model. The PRC's second variable is precision, the proportion of identified hits that are actually hits, whereas the ROC's second variable is the false positive rate, the percentage of hits incorrectly identified as misses. The precision is more relevant to virtual screening, which is focused on maximizing the ratio of hits to misses in a small sample, rather than minimizing the number of hits that were mislabeled. An exception was made for CoV-2 3CL and CoV-2 ChemProp3D/ChemProp3D-NDU. Here the models learned slowly, and early epochs contained high PRC values, but also ROC values below 0.5 (worse than a random model). Hence the ROC was a more informative metric for hyperparameter optimization.

Random forest models were trained as regressors rather than classifiers, as the ROC and PRC scores do not make sense when the predictions themselves are binary. Any predictions greater than 1 were automatically set to 1, and any predictions less than 0 to 0, though all model predictions were already in the range (0, 1). Hyperparameters were also optimized with Bayesian methods \cite{hyperopt}. We used the ROC score on the validation set to evaluate hyperparameters, because it consistently gave the most robust models (i.e. those with both high ROC and PRC scores). Ten models were trained with different initial seeds, and balanced sampling was used during training. Details of the hyperparameter ranges can be found in \cite{nff}, and further details about the optimization are in \cite{dataverse_models}.

\section{Scoring metrics}

In the ML community, the standard metric for evaluating binary classifiers is the receiver operating characteristic area under the curve (ROC-AUC) \cite{Wu2018MoleculeNet, chemprop, mayr2018large}. The ROC curve plots the true positive rate (TPR) on the $y$-axis against the false positive rate (FPR) on the $x$-axis. The plot is created by considering different thresholds for which a predicted probability is considered a hit. For example, a threshold of 0 gives an FPR of 1 and a TPR of 1, while a threshold of 1 gives an FPR of 0 and a TPR of 0. By varying the threshold one arrives at a set of (FPR, TPR) pairs. The ROC-AUC is the area under the FPR-TPR curve; it is 0.5 for a random model and 1.0 for a perfect model. 

An alternative to the ROC curve is the precision-recall curve (PRC), which plots the precision on the $y$-axis (the proportion of predicted hits that are actually hits) against recall on the $x$-axis (same as TPR). For a perfect model the PRC-AUC is equal to 1.0, while for a random model it is equal to the proportion of hits in the dataset. The dependence on the makeup of the dataset is a drawback, since, unlike the ROC-AUC, the PRC-AUC is not transferable among models tested on different datasets. However, its use of precision makes it more applicable to the early retrieval problem in drug discovery. Moreover, when the dataset is imbalanced, which is almost always the case in drug discovery, the PRC-AUC is strictly more informative than the ROC-AUC \cite{prc_auc_better}.

A significant problem with both metrics is that they compute the area under the \textit{entire} curve. This is problematic because high $x$ values correspond to low thresholds, i.e. thresholds for which most species are considered hits. The purpose of virtual screening is to select the top few candidates for experimental testing. One would never use a threshold for which a majority of candidates are considered hits, because then one would have to test a large portion of the sample. If one could test a large portion then, presumably, one could test \textit{all} candidates, and so virtual screening would be unnecessary. While this point has long been understood in the computational drug discovery community \cite{jain2008recommendations}, it has largely been overlooked in the deep learning literature (with the notable exception of Ref. \cite{feinberg2018potentialnet}).

An alternative is the enrichment factor, defined as $\mathrm{TP}_{\mathrm{model}} / \mathrm{TP}_{\mathrm{rand}}$. Here $\mathrm{TP}_{\mathrm{model}}$ is the number of true hits among species in the top few percent of predictions, and $\mathrm{TP}_{\mathrm{rand}}$ is the number that would be retrieved by a random model \cite{jain2008recommendations}. However, like the PRC-AUC it too depends on the ratio of hits to misses in the dataset \cite{nicholls2008we}. To address this issue, and to standardize the definition of ``top few percent'', Ref. \cite{jain2008recommendations} recommends using the so-called ROC enrichment (ROCE), defined as the ratio of the $y$ value to the $x$ value on an ROC plot, at the $x$ values of 0.5\%, 1\%, 2\%, and 5\%. In this work we reported the ROCE values in addition to the ROC-AUC and PRC-AUC scores, and suggest that future works in deep learning for drug discovery do the same.
 
\section{Uncertainty quantification}
For less intensive models like ChemProp with small amounts of data, test score uncertainty was determined by training ten different models and calculating the standard deviation in their scores. For evaluating the test set PRC, the model with the best validation PRC was used, and similarly for the ROC. For the ROCE scores the average was taken over all 20 models (ten models with the best validation PRC and ten with the best validation ROC). 

The ensemble method of uncertainty quantification was not feasible for any of the 3D models, or for ChemProp trained on the CoV 3CL data. In these cases we used two different models from different epochs in the same training progression. Each model had the best validation score according to either the PRC or ROC metric. Averages and standard deviations were then calculated from two scores, one from each of the two models evaluated on the test set.  This approach is justified because, in many cases, the model with the best validation PRC/ROC was not the one with the best test PRC/ROC. Note, however, that this method gives zero uncertainty whenever the epoch with the best ROC also has the best PRC. Further details of the training process can be found in the SM, and exact parameters used for each of the different models can be found in \cite{dataverse_models}.

\bibliographystyle{unsrtnat}
\bibliography{main}

\end{document}